\documentclass[runningheads]{llncs}
\usepackage{graphicx}

\usepackage{tikz}
\usepackage{comment}
\usepackage{amsmath,amssymb} 
\usepackage{color}

\usepackage[accsupp]{axessibility}  

\usepackage{hyperref}
\hypersetup{
    colorlinks=true,
    linkcolor=blue,
    filecolor=magenta,      
    urlcolor=cyan,
    }
    
\usepackage{color,xcolor}
\usepackage{epsfig}
\usepackage{graphicx}

\usepackage{adjustbox}
\usepackage{array}
\usepackage{booktabs}
\usepackage{colortbl}
\usepackage{float,wrapfig}
\usepackage{hhline}
\usepackage{multirow}
\usepackage{caption}

\usepackage{cleveref}

\usepackage{bm}
\usepackage{nicefrac}
\usepackage{mathtools}

\usepackage{changepage}
\usepackage{extramarks}
\usepackage{fancyhdr}
\usepackage{lastpage}
\usepackage{setspace}
\usepackage{xspace}

\usepackage{url}

\usepackage{amssymb}
\usepackage{amsfonts}
\usepackage{algorithmic}
\usepackage{algorithm}
\usepackage{todonotes} 
\usepackage{enumitem}
\usepackage{bbm}

\makeatletter
\DeclareRobustCommand\onedot{\futurelet\@let@token\@onedot}
\def\@onedot{\ifx\@let@token.\else.\null\fi\xspace}

\def\eg{\emph{e.g}\onedot} 
\def\ie{\emph{i.e}\onedot}

\def\etal{\emph{et al}\onedot}
\makeatother

\def\N{{\mathcal N}}

\newcommand{\eqn}[1]{Equation~(\ref{#1})}
\newcommand{\fig}[1]{Figure~\ref{#1}}
\newcommand{\tbl}[1]{Table~\ref{#1}}

\newcommand{\ignore}[1]{}

\definecolor{rowblue}{RGB}{220,230,240}
\definecolor{myorchid}{RGB}{150,10,30}
\definecolor{myblue}{RGB}{10,30,250}
\definecolor{mygreen}{RGB}{10,120,10}

\usepackage{amsmath,amsfonts,bm}

\def\eqref#1{equation~\ref{#1}}

\def\1{\bm{1}}

\newcommand{\test}{\mathcal{D_{\mathrm{test}}}}

\def\vc{{\bm{c}}}

\def\vx{{\bm{x}}}

\DeclareMathAlphabet{\mathsfit}{\encodingdefault}{\sfdefault}{m}{sl}
\SetMathAlphabet{\mathsfit}{bold}{\encodingdefault}{\sfdefault}{bx}{n}

\newcommand{\R}{\mathbb{R}}

\usepackage{orcidlink}
\usepackage[width=122mm,left=12mm,paperwidth=146mm,height=193mm,top=12mm,paperheight=217mm]{geometry}
\begin{document}
\pagestyle{headings}
\mainmatter
\def\ECCVSubNumber{100}  

\title{Compositional Visual Generation with \\ Composable Diffusion Models} 

\titlerunning{Composable Diffusion}
%
\author{Nan Liu\inst{1}\thanks{indicates equal contribution. \\ Correspondence to: Shuang Li $<$lishuang@mit.edu$>$, Yilun Du $<$yilundu@mit.edu$>$}\orcidlink{0000-0002-1380-5428} \and
Shuang Li\inst{2 \star}\orcidlink{0000-0002-7276-5032}  \and
Yilun Du\inst{2 \star}\orcidlink{0000-0001-6792-5946}  \\
Antonio Torralba\inst{2}  \and
Joshua B. Tenenbaum\inst{2}
}
\authorrunning{Nan Liu*, Shuang Li*, Yilun Du* (equal contribution)}
%
\institute{University of Illinois Urbana-Champaign \and
Massachusetts Institute of Technology \\
\email{nanliu4@illinois.edu, \{lishuang,yilundu,torralba,jbt\}@mit.edu}}
\maketitle

\begin{abstract}
Large text-guided diffusion models, such as DALL-E 2, are able to generate stunning photorealistic images given natural language descriptions. While such models are highly flexible, they struggle to understand the composition of certain concepts, such as confusing the attributes of different objects or relations between objects. In this paper, we propose an alternative structured approach for compositional generation using diffusion models. An image is generated by composing a set of diffusion models, with each of them modeling a certain component of the image. To do this, we interpret diffusion models as energy-based models in which the data distributions defined by the energy functions may be explicitly combined. The proposed method can generate scenes at test time that are substantially more complex than those seen in training, composing sentence descriptions, object relations, human facial attributes, and even generalizing to new combinations that are rarely seen in the real world. We further illustrate how our approach may be used to compose pre-trained text-guided diffusion models and generate photorealistic images containing all the details described in the input descriptions, including the binding of certain object attributes that have been shown difficult for DALL-E 2. These results point to the effectiveness of the proposed method in promoting structured generalization for visual generation. 
\let\thefootnote\relax\footnotetext{Webpage: \href{https://energy-based-model.github.io/Compositional-Visual-Generation-with-Composable-Diffusion-Models/}{\small{https://energy-based-model.github.io/Compositional-Visual-Generation-with-Composable-Diffusion-Models/}}}
\end{abstract}

\section{Introduction}

Our understanding of the world is highly compositional in nature. 
We are able to rapidly understand new objects from their components, or compose words into complex sentences to describe the world states we encounter \cite{lakehuman}.  We are able to make ``infinite use of finite means'' \cite{chomsky1965}, \ie, repeatedly reuse and recombine concepts we have acquired in a potentially infinite manner. We are interested in constructing machine learning systems to have such compositional capabilities, particularly in the context of generative modeling. 

\begin{figure}[t]
\begin{center}
\includegraphics[width=1\textwidth]{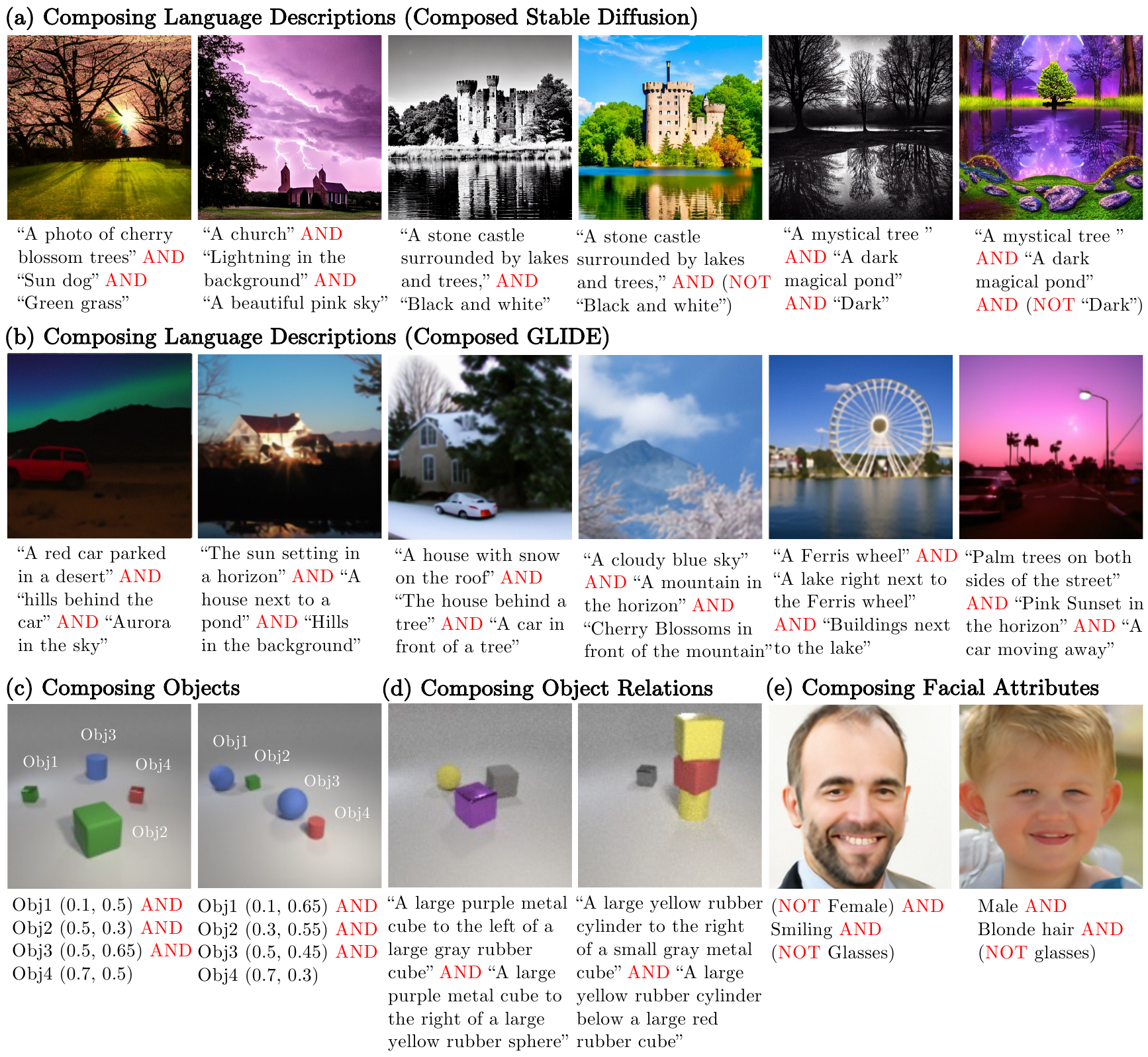}
\end{center}
\vspace{-15pt}
\caption{\small Our method allows compositional visual generation across a variety of domains, such as language descriptions, objects, object relations, and human attributes.}
\label{fig:teaser}
\vspace{-15pt}
\end{figure}

Existing text-conditioned diffusion models such as DALL-E 2 \cite{dalle2} have recently made remarkable strides towards compositional generation, and are capable of generating photorealistic images given textual descriptions. However, such systems are not fully compositional and generate incorrect images when given more complex descriptions \cite{marcus2022very,swimmer963_2022}. An underlying difficulty may be that such models encode text descriptions as fixed-size latent vectors. 
However, as textual descriptions become more complex, more information needs to be squeezed into the fixed-size vector. 
Thus it is impossible to encode arbitrarily complex textual descriptions.

In this work, we propose to factorize the compositional generation problem, using different diffusion models to capture different subsets of a compositional specification. These diffusion models are then explicitly composed together to generate an image. By explicitly factorizing the compositional generative modeling problem, our method can generalize to significantly more complex combinations that are unseen during training.

Such an explicit form of compositionality has been explored before under the context of Energy-Based Models (EBMs) \cite{du2020compositional,du2021unsupervised,liu2021learning}. 
However, directly training EBMs has been proved to be unstable and hard to scale.
We show that diffusion models can be interpreted as implicitly parameterized EBMs, which can be further composed for image generation, significantly improving training stability and image quality.

Our proposed method enables zero-shot compositional generation across different domains as shown in \fig{fig:teaser}. First, we illustrate how our approach may be applied to large pre-trained diffusion models, such as Stable Diffusion~\cite{rombach2021highresolution} and GLIDE~\cite{nichol2021glide}, to compose multiple text descriptions.
Next, we illustrate how our approach can be applied to compose objects and object relations, enabling zero-shot generalization to a larger number of objects. Finally, we illustrate how our framework can compose different facial attributes to generate human faces.

\textbf{Contributions:} In this paper, we introduce an approach towards compositional visual generation using diffusion models. First, we show that diffusion models can be composed by interpreting them as energy-based models, and drawing on this connection, we demonstrate how to compose diffusion models together. Second, we propose two compositional operators, \emph{Conjunction} and \emph{Negation}, on top of diffusion models that allow us to compose concepts in different domains during inference without any additional training. We show that the proposed method enables effective zero-shot combinatorial generalization, \ie generating images with more complicated compositions of concepts. Finally, we evaluate our method on composing language descriptions, objects, object relations, and human facial attributes. Our method can generate high-quality images containing all the concepts and outperforms baselines by a large margin. For example, the accuracy of our method is $24.02\%$ higher than the best baseline for composing three objects in specified positions on the CLEVR dataset.

\section{Related Work}
\label{related_work}
\noindent\textbf{Controllable Image Generation.} Our work is related to existing work on controllable image generation. One type of approach towards controllable image generation specifies the underlying content of an image utilizing text through GANs \cite{xu2018attngan,zhang2017stackgan,bau2021paint}, VQ-VAEs \cite{ramesh2021zero}, or diffusion models \cite{nichol2021glide}. An alternative type of approach towards controllable image generation manipulates the underlying attributes in an image \cite{shoshan2021gan,xiao2018elegant,zhu2020domain}. In contrast, we are interested in \textit{compositionally controlling} the underlying content of an image at test time, generating images that exhibit compositions of multiple types of image content. Thus, most relevant to our work, existing work has utilized EBMs to compose different factors describing a scene \cite{du2020compositional,lace,du2021unsupervised,liu2021learning}. We illustrate how we may implement such probabilistic composition on diffusion models, achieving better performance. 

\noindent\textbf{Diffusion Models.} Diffusion models have emerged as a promising class of generative models that formulates the data-generating process as an iterative denoising procedure \cite{sohl2015deep,ho2020denoising}. The denoising procedure can be seen as parameterizing the gradients of the data distribution \cite{song2020score}, which is similar to EBMs \cite{lecun2006tutorial,du2019implicit,nijkamp2019anatomy,grathwohl2020stein,gao2021learning}. Diffusion models have recently shown great promise in image generation tasks \cite{dhariwal2021diffusion}, enabling effective image editing \cite{meng2021sdedit,kim2021diffusionclip}, text conditioning \cite{nichol2021glide,rombach2022high,gu2022vector}, and image inpainting \cite{saharia2021palette}. The iterative, gradient-based sampling of diffusion models enable us to compose multiple factors during inference. While diffusion models have been developed for image generation \cite{song2020denoising}, they have further proven successful in the generation of waveforms \cite{chen2020wavegrad}, 3D shapes \cite{zhou2021shape}, decision making \cite{janner2022diffuser}, and text \cite{austin2021structured}, suggesting that our proposed composition operators may further be applied in such domains.
\section{Background}

\subsection{Denoising Diffusion Models}
\label{diffusion}
Denoising Diffusion Probabilistic Models (DDPMs) are a class of generative models where generation is modeled as a denoising process. Starting from a sampled noise, the diffusion model performs $T$ denoising steps until a sharp image is formed.  In particular, the denoising process produces a series of intermediate images with decreasing levels of noise, denoted as $\vx_T, \vx_{T-1}, ..., \vx_0$, where $\vx_T$ is sampled from a Gaussian prior and $\vx_0$ is the final output image. 

DDPMs construct a forward diffusion process by gradually adding Gaussian noise to the ground truth image. A diffusion model then learns to revert this noise corruption process. Both the \textit{forward processes} $q(\vx_t | \vx_{t-1})$ and the \textit{reverse process} $p_{\theta}(\vx_{t-1}|\vx_t)$ are modeled as the products of Markov transition probabilities: 
\begin{equation}
\label{eq:factorize}
    \begin{aligned}
        q(\vx_{0:T}) &= q(\vx_0)\prod_{t=1}^{T} q(\vx_t | \vx_{t-1}), \;\;\;\;\;
    p_\theta(\vx_{T:0}) =  p(\vx_T)\prod_{t=T}^{1} p_\theta(\vx_{t-1} | \vx_t),
    \end{aligned}
\end{equation}
where $q(\vx_0)$ is the real data distribution and $p(\vx_T)$ is a standard Gaussian prior. 

A \textit{generative process} $p_\theta(\vx_{t-1} | \vx_t)$ is trained to generate realistic images by approximating the reverse process through variational inference. Each step of the \textit{generative process} is a Gaussian distribution $\N$ with a learned mean $\mu_\theta(\vx_t, t)$ and covariance matrix $\sigma_t^2 I$, where $I$ is the identity matrix.
\begin{equation}
\begin{aligned}
    p_\theta( \vx_{t-1}|\vx_t) &:= \N \bigl(\mu_\theta(\vx_t, t), \sigma_t^2 I \bigl) = \N \bigl(\vx_{t} - \epsilon_\theta(\vx_t, t \bigl), \sigma_t^2 I).
\end{aligned}
\end{equation}
The mean $\mu_\theta(\vx_t, t)$ is represented by a perturbation $\epsilon_\theta(\vx_t, t)$ to a noisy image $\vx_t$.
The goal is to remove the noise gradually by predicting a less noisy image at timestep $\vx_{t-1}$ given a noisy image $\vx_t$. 
To generate real images, we sample $\vx_{t-1}$ from $t=T$ to $t=1$ using the parameterized marginal distribution $ p_\theta( \vx_{t-1}|\vx_t)$, with an individual step corresponding to:
\begin{equation}
   \vx_{t-1} = \vx_t - \epsilon_\theta(\vx_t, t) + \N(0, \sigma_t^2 I).
   \label{eqn:diffusion_langevin}
\end{equation}
The generated images become more realistic over multiple iterations.

\subsection{Energy Based Models}
\label{ebm}
Energy-Based Models (EBMs) \cite{du2019implicit,du2020improved,grathwohl2020stein,nijkamp2019anatomy} are a class of generative models where the data distribution is modeled using an unnormalized probability density. Given an image $\vx \in \mathbb{R}^D$, the probability density of image $\vx$ is defined as:
\begin{equation}
    \label{eq:ebm}
    p_{\theta}(\vx) \propto e^{-E_{\theta}(\vx)},
\end{equation}
where the energy function $E_\theta(\vx): \R^D \rightarrow \R$ is a learnable neural network.
A gradient based MCMC procedure, Langevin dynamics \cite{du2019implicit}, is then used to sample from the unnormalized probability distribution to iteratively refine the generated image $\vx$: 
\begin{equation}
    \label{eq:langevin}
    \vx_t = 
    \vx_{t - 1} - 
    \frac{\lambda}{2}\nabla_{\vx} E_{\theta}(\vx_{t-1}) +
    \mathcal{N}(0, \sigma_t^2 I).
\end{equation}
The sampling procedure used by diffusion models in \eqn{eqn:diffusion_langevin} is functionally similar to the sampling procedure used by EBMs in \eqn{eq:langevin}. In both settings, images are iteratively refined starting from a Gaussian noise, with a small amount of additional noise added at each iterative step.

\begin{figure}[t]
\begin{center}
\includegraphics[width=1\textwidth]{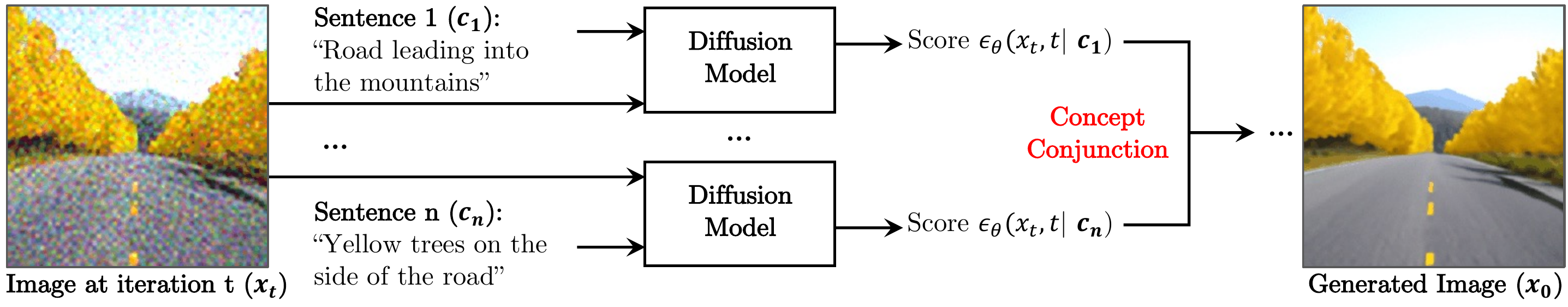}
\end{center}
\vspace{-10pt}
\caption{\small \textbf{Compositional generation.} Our method can compose multiple concepts during inference and generate images containing all the concepts without further training. We first send an image from iteration $t$ and each of the concepts to the diffusion model to generate a set of scores $\{\epsilon_\theta(\vx_t, t|\vc_1), \ldots, \epsilon_\theta(\vx_t, t|\vc_n)\}$. We then compose different concepts using the proposed compositional operators, such as conjunction, to denoise the generated image. The final image is obtained after $T$ iterations.}
\label{fig:model}
\vspace{-10pt}
\end{figure}

\section{Our approach}
In this section, we first introduce how we interpret diffusion models as energy-based models in \cref{gradient_ebm} and then introduce how we compose diffusion models for visual generation in \cref{sec:compositionality_diffusion}.

\subsection{Diffusion Models as Energy Based Models}
\label{gradient_ebm}
The sampling procedure of diffusion models in \eqn{eqn:diffusion_langevin} and EBMs in \eqn{eq:langevin} are functionally similar.
At a timestep $t$, in diffusion models, images are updated using a learned denoising network $\epsilon_\theta(\vx_t, t)$ while in EBMs, images are updated using the gradient of the energy function $\nabla_{\vx} E_\theta(\vx_t) \propto \nabla_{\vx} \log p_\theta(\vx_t)$.
The denoising network $\epsilon_\theta(\vx_t, t)$ is trained to predict the underlying score of the data distribution \cite{vincent2011connection,song2020denoising} when the number of diffusion steps increases to infinity.  Similarly, an EBM is trained so that $\nabla_\vx E_\theta(\vx_t)$ corresponds to the score of the data distribution as well. In this sense, $\epsilon_\theta(\vx_t, t)$ and $\nabla_\vx E_\theta(\vx_t)$ are functionally the same, and the underlying sampling procedure in \eqn{eqn:diffusion_langevin} and \eqn{eq:langevin} are equivalent. We may view a trained diffusion model $\epsilon_\theta(\vx_t, t)$ as an implicitly parameterized EBM. Such parameterization enables us to apply previous techniques for composing EBMs to diffusion models.

\vspace{+5pt}
\noindent\textbf{Composing EBMs.} Previous EBMs \cite{hinton2002training,du2020compositional,liu2021learning} have shown good performance on compositional visual generation.
Given $n$ independent EBMs, $E^1_\theta(\vx), \cdots, E^n_\theta(\vx)$, the functional form of EBMs in \eqn{eq:ebm} enables us to compose multiple separate EBMs together to obtain a new EBM. The composed distribution can be represented as:
\begin{equation}
    \label{eq:1}
    p_{\text{compose}}(\vx) \propto p_{\theta}^1(\vx) \cdots p_{\theta}^n(\vx) \propto e^{- \sum_{i=1}^{n} E_{\theta}^i(\vx)  }, 
\end{equation}
where $p_{\theta}^i \propto e^{-E_{\theta}^i(\vx)}$ is the probability density of image $\vx$ (\eqn{eq:ebm}).
Langevin dynamics is then used to iteratively refine the generated image $\vx$~\cite{du2020compositional,liu2021learning}.

\begin{equation}
    \label{eq:langevin_condition}
    \vx_t = 
    \vx_{t - 1} - 
    \frac{\lambda}{2}\nabla_{\vx} \left(\sum_{i=1}^n E_{\theta}^i(\vx_{t-1}) \right) +
    \mathcal{N}(0, \sigma_t^2 I).
\end{equation}

\noindent\textbf{Composing Diffusion Models.}
By leveraging the interpretation that diffusion models are functionally similar to EBMs, we may compose diffusion models in a similar way. 
The \textit{generative process} and the score function of a diffusion model can be represented as $p^i_\theta(\vx_{t-1}|\vx_{t})$ and $\epsilon^i_\theta(\vx_t, t)$, respectively.
If we treat the score functions in diffusion models as the learned gradient of energy functions in EBMs, the score function of the composed diffusion model can be written as $\sum_{i=1}^n \epsilon^i_\theta(\vx_t, t)$.
Thus the \textit{generative process} of composing multiple diffusion models becomes:
\begin{equation}
    \label{eqn:diffusion_compose}
    p_{\text{compose}}( \vx_{t-1}|\vx_t) = \N \left(\vx_t - \sum_{i=1}^n \epsilon^i_\theta(\vx_t, t), \sigma_t^2 I \right).
\end{equation}

A complication of parameterizing a gradient field of EBM $\nabla_\vx E_\theta(\vx_t)$ with a learned score function $\epsilon_\theta(\vx_t, t)$ is that the gradient field may not be conservative, and thus does not correspond to a valid probability density. However, as discussed in \cite{salimans2021should}, explicitly parameterizing the learned function $\epsilon_\theta(\vx_t, t)$ as the gradient of EBM achieves similar performance as the non-conservative parameterization of diffusion models, suggesting this is not problematic.

\subsection{Compositional Generation through Diffusion Models}
\label{sec:compositionality_diffusion}
Next, we discuss how we compose diffusion models for image generation. We aim to generate images conditioned on a set of concepts $\{\vc_1, \vc_2, \ldots, \vc_n\}$. To do this, we represent each concept $\vc_i$ using a diffusion model, which can be composed to generate images.
Inspired by EBMs~\cite{du2020compositional,liu2021learning}, we define two compositional operators, \textbf{conjunction (AND)} and \textbf{negation (NOT)}, to compose diffusion models.
We learn a set of diffusion models representing the conditional probability distribution $p(\vx|\vc_i)$ given concept $\vc_i$ and an unconditional probability distribution $p(\vx)$.

\vspace{+5pt}
\noindent\textbf{Concept Conjunction (AND).} We aim to generate images containing certain attributes.
Following \cite{du2020compositional}, the conditional probability can be factorized as:
\begin{equation}
   p(\vx|\vc_1, \ldots, \vc_n) \propto p(\vx, \vc_1, \ldots, \vc_n) = p(\vx) \prod_{i=1}^n p(\vc_i|\vx).
   \label{eq:conj}
\end{equation}
Here we assume the concepts are conditionally independent given $\vx$.
We can represent $p(\vc_i|\vx)$ using the combination of a conditional distribution $p(\vx|\vc_i)$ and an unconditional distribution $p(\vx)$, with both of them are parameterized as diffusion models $p(\vc_i|\vx) \propto \frac{p(\vx|\vc_i)}{p(\vx)}$.
The expression of $p(\vc_i|\vx)$ corresponds to the implicit classifier that represents the likelihood of $\vx$ exhibiting concept $\vc_i$.
Substituting $p(\vc_i|\vx)$ in Equation \ref{eq:conj}, we can rewrite Equation \ref{eq:conj} as:
\begin{equation}
     p(\vx|\vc_1, \ldots, \vc_n) \propto p(\vx) \prod_{i=1}^n \frac{p(\vx|\vc_i)}{p(\vx)}.
     \label{eq:conj_works}
\end{equation}
We sample from this resultant distribution using \eqn{eqn:diffusion_compose} with the composed score function $\hat{\epsilon}(\vx_t, t)$:
\begin{equation}
    \hat{\epsilon}(\vx_t, t) = \epsilon_\theta(\vx_t, t) + \sum_{i=1}^n w_i \bigl(\epsilon_\theta(\vx_t, t | \vc_i) - \epsilon_{\theta}(\vx_t, t)\bigl),
    \vspace{-5pt}
\end{equation}
where $w_i$ is a hyperparameter corresponding to the temperature scaling on concept $\vc_i$. 
We can generate images with the composed concepts using the following \textit{generative process}:
\begin{equation}
    p_{compose}( \vx_{t-1}|\vx_t) := \N \bigl(\vx_t - \hat{\epsilon}(\vx_t, t), \sigma_t^2 I \bigl).
    \label{eqn:diffusion_mod}
\end{equation}
In the setting in which image generation is conditioned on a single concept, the above sampling procedure reduces to the classifier-free guidance~\cite{ho2021classifier}.

\begin{figure}[t]
\centering
\small
\vspace{-10pt}
\scalebox{1}{
\begin{minipage}{\linewidth}
    \begin{algorithm}[H]
        \vspace{-2pt}
        \algsetup{linenosize=\small}
        \caption{Code for Composing Diffusion Models}
        \label{code}
        \begin{algorithmic}[1]
        \STATE \textbf{Require} Diffusion model $\epsilon_\theta(\vx_t, t|\vc)$, scales $w_i$ and $w$, covariance matrix $\sigma_t^2 I$ \\
        \STATE \small{\color{gray}// Code for conjunction}
        \STATE Initialize sample  $\vx_T \sim \mathcal{N}(\bm{0}, \bm{I})$  \\
        \FOR{$t = T, \ldots, 1$}
            \STATE $\epsilon_i \gets \epsilon_\theta(\vx_{t}, t | \vc_i)$ \hspace{1.8cm} \small{\color{gray}// compute conditional scores for each concept $\vc_i$} \\
            \STATE $\epsilon \gets \epsilon_\theta(\vx_{t}, t)$ \hspace{4.8cm} \small{\color{gray}// compute unconditional score}  \\
            \STATE $\vx_{t-1} \sim \mathcal{N} \Bigl(\vx_t - \bigl(\epsilon +  \sum_{i=1}^n w_i (\epsilon_i - \epsilon)\bigl), \sigma_t^2 I \Bigl)$ \hspace{3.5cm} \small{\color{gray}// sampling} \\
        \ENDFOR \\
        \STATE 
        \STATE \small{\color{gray}// Code for negation}
        \STATE Initialize sample  $\vx_T \sim \mathcal{N}(\bm{0}, \bm{I})$ \\
        \FOR{$t = T, \ldots, 1$}
            \STATE $\tilde{\epsilon}_j \gets \epsilon_\theta(\vx_{t}, t | \tilde{\vc}_j)$  \hspace{0.8cm} \small{\color{gray}// compute conditional score for the negated concept $\tilde{\vc}_j$} \\
            \STATE $\epsilon_i \gets \epsilon_\theta(\vx_{t}, t | \vc_i)$  \hspace{2.6cm} \small{\color{gray}// compute conditional score for concept $\vc_i$} \\
            \STATE $\epsilon \gets \epsilon_\theta(\vx_{t}, t)$ \hspace{4.8cm} \small{\color{gray}// compute unconditional score} \\
            \STATE $\vx_{t-1} \sim \mathcal{N}\Bigl(\vx_t - \bigl(\epsilon + w (\epsilon_i - \tilde{\epsilon}_j)\bigl), \sigma_t^2 I\Bigl)$ \hspace{4.1cm} \small{\color{gray}// sampling}\\
        \ENDFOR \\
        \end{algorithmic}
    \end{algorithm}
\end{minipage}
}
\vspace{-15pt}
\end{figure}

\vspace{+5pt}
\noindent\textbf{Concept Negation (NOT).} In concept negation, we aim to generate realistic images with the absence of a certain factor $\tilde{\vc}_j$. 
However, the negation of a concept can be ill-defined. For example, the negation of ``dark'' can be ``bright'' or random noises.
Thus we generate images conditioned other concepts as well to make the generated images look real.
Following \cite{du2020compositional}, concept negation can be represented as the composed probability distribution $p(\vx|\text{not } \tilde{\vc}_j, \vc_i)$.
Similarly, we refactorize the joint probability distribution as:
\begin{equation}
  p(\vx|\text{not }\tilde{\vc}_j, \vc_i) \propto p(\vx, \text{not }\tilde{\vc}_j, \vc_i) \propto p(\vx) \frac{ p(\vc_i|\vx)}{p(\tilde{\vc}_j|\vx)}.
  \label{eqn:negation_first}
\end{equation}
Using the factorization $p(\vc_i|\vx) \propto \frac{p(\vx|\vc_i)}{p(\vx)}$, we can rewrite \eqn{eqn:negation_first} as:

\begin{equation}
    p(\vx|\text{not }\tilde{\vc}_j, \vc_i) \propto p(\vx)\frac{p(\vx | \vc_i)}{p(\vx | \tilde{\vc}_j)}
\end{equation}
We may construct the composed score funcion $\hat{\epsilon}(\vx, t)$ as:
\begin{equation}
    \small
    \hat{\epsilon}(\vx_t, t) = \epsilon_\theta(\vx_t, t) + w \bigl(\epsilon_\theta(\vx_t, t | \vc_i) - \epsilon_\theta(\vx_t, t | \tilde{\vc}_j)\bigl).
\end{equation}
where $w$ is the hyperparameter that controls the strength of the negation. We can generate samples using this composed score function and Equation \ref{eqn:diffusion_mod}. 
Algorithm~\ref{code} provides the pseudo-code for composing diffusion models using concept conjunction and negation. 
Our method can compose pre-trained diffusion models during inference without any additional training. Please see the full derivation details for both operators in~\cref{sup:proof}.

\section{Experiment Setup}
\subsection{Datasets}
\label{datasets}

\textbf{CLEVR.}
CLEVR~\cite{johnson2017clevr} is a synthetic dataset containing objects with different shapes, colors, and sizes.
The training set consists of 30,000 images at $128 \times 128$ resolution. Each image contains $1 \sim 5$ objects and a 2D coordinate $(x, y)$ label indicating that the image contains an object at $(x, y)$. In our experiments, the 2D coordinate label is the coordinate of one object in the image. 

\noindent\textbf{Relational CLEVR.}
Relational CLEVR~\cite{liu2021learning} contains relational descriptions between objects in the image, such as ``a red cube to the left of a blue cylinder''.
The training dataset contains $50,000$ images at $128 \times 128$ resolution. Each training image contains $1 \sim 5$ objects and one label describing a relation between two objects. If there is only one object in the image, the second object and their relation in the relational description are both nulls.

\noindent\textbf{FFHQ.}
FFHQ~\cite{karras2019style} is a real-world human face dataset. The original FFHQ dataset consists of 70,000 human face images without labels.
\cite{ffhq-features-dataset} annotates three binary attributes, including \textit{smile}, \textit{gender}, and \textit{glasses}, for the images using pre-trained classifiers. In total, there are 51,067 images labeled by the classifiers. 

\subsection{Evaluation Metrics}
\label{exp_evaluation_metrics}
\textbf{Binary classification accuracy.}
During testing, we evaluate the performance of the proposed method and baselines on three different settings.
The first test setting, \textbf{1 Component}, generates images conditioned on a single concept (matching the training distribution). The second and third test settings, \textbf{2 Components} and \textbf{3 Components}, generate images by composing two and three concepts, respectively, using the \emph{conjunction} and \emph{negation} operators. They are used to evaluate the models' generalization ability to new combinations. 

For each task, we use the training data (real images) to train a binary classifier that takes an image and a concept, \eg `smiling', as input, and predicts whether the image contains or represents the concept. We then apply this classifier to a generated image, checking whether it faithfully captures each of the concepts.
In each test setting, each method generates $5,000$ images for evaluation. The accuracy of the method is the percentage of generated images capturing all the concepts 
(See \cref{sup:binary_classification}).

\noindent\textbf{Fréchet Inception Distance (FID)} is a commonly used metric for evaluating the quality of generated images. It uses a pre-trained inception model~\cite{inception} to extract features for the generated images and real images, and measures their feature similarity. Specifically, we use Clean-FID~\cite{parmar2021cleanfid} to evaluate the generated images. FID is usually computed on $50,000$ generated images, but we use $5,000$ images in our experiments.

\section{Experiments}
\label{exp:compose_different_concepts}

We compare the proposed method and baselines (\cref{sec:baselines}) on compositional generation in different domains.
We show results of composing natural language descriptions (\cref{exp:compose_language}), objects (\cref{exp:compose_object_attributes}), object relational descriptions (\cref{exp:compose_object_relations}), and human facial attributes (\cref{sup:additional_results}).
Results analysis are shown in \cref{exp:result_analysis}.

\begin{figure}[t]
\begin{center}
\includegraphics[width=1\textwidth]{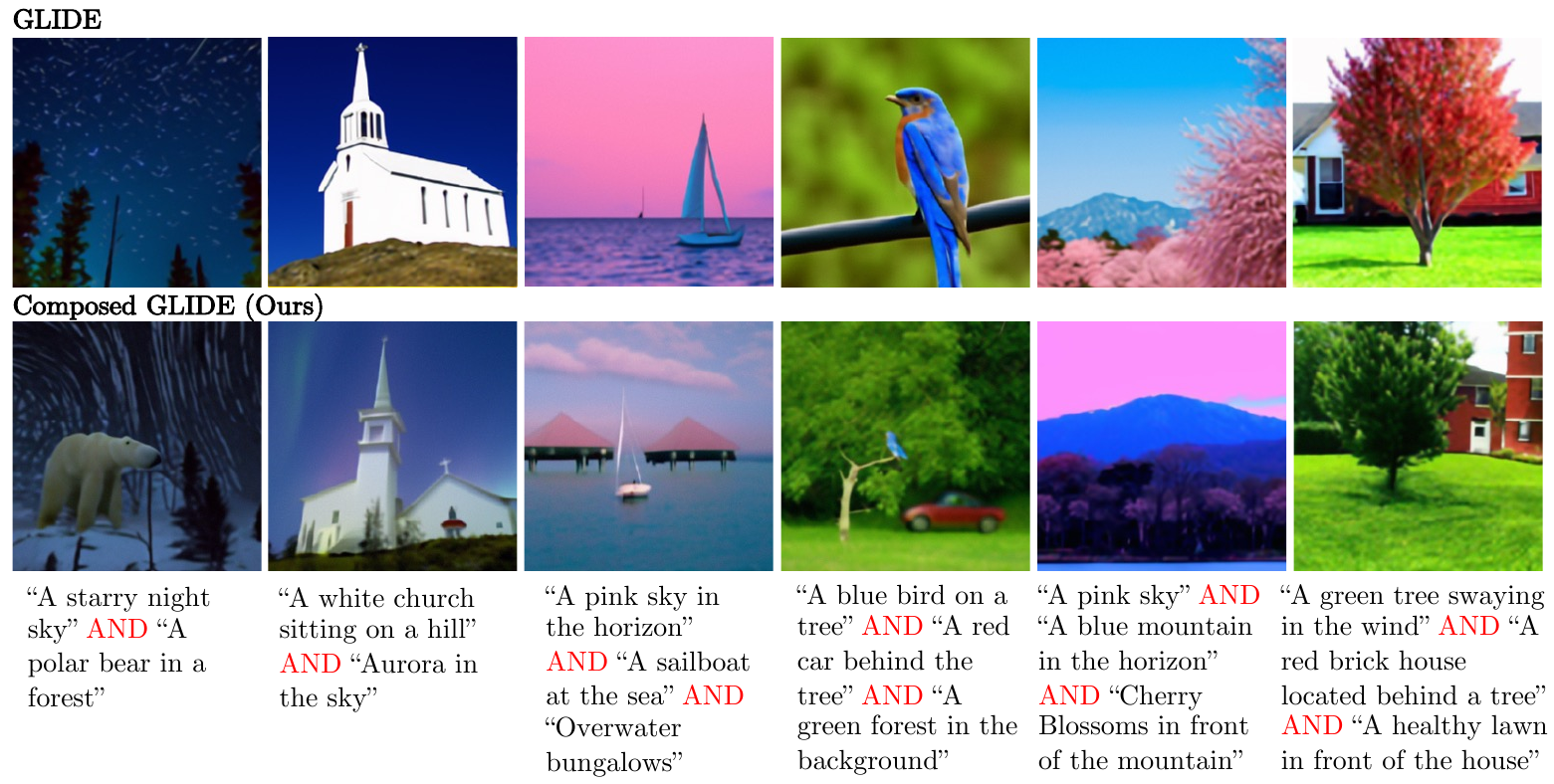}
\end{center}
\vspace{-15pt}
\caption{\small \textbf{Composing Language Descriptions.} We develop Composed GLIDE (Ours), a version of GLIDE~\cite{nichol2021glide} that utilizes our compositional operators to combine textual descriptions, without further training. We compare it to the original GLIDE, which directly encodes the descriptions as a single long sentence. 
Our approach more accurately captures text details, such as the ``overwater bungalows'' in the third example.}
\label{fig:image_generation_glide}
\vspace{-10pt}
\end{figure}

\subsection{Baselines}
\label{sec:baselines}

We compare our method with baselines for compositional visual generation.

\noindent\textbf{StyleGAN2-ADA~\cite{stylegan2ada}} is the state-of-the-art GAN method for both unconditional and conditional image generation.

\noindent\textbf{StyleGAN2~\cite{stylegan2}} is one of the state-of-the-art GAN methods for unconditional image generation. To enable compositional image generation, 
we optimize the latent code $z$ by decreasing the binary classification loss of the generated image and the given label. We use the resultant latent code to generate images.

\noindent\textbf{LACE~\cite{lace}} uses pre-trained classifiers to generate energy scores in the latent space of the pre-trained StyleGAN2 model. To enable compositional image synthesis, LACE uses compositional operators~\cite{du2020compositional}.

\noindent\textbf{GLIDE~\cite{nichol2021glide}} is a recently released text-conditioned diffusion model for image generation. 
For composing language descriptions, we use the pre-trained GLIDE released by OpenAI. For the rest tasks, we use the GLIDE code and train a model on each task.

\noindent\textbf{Energy-based models (EBM)~\cite{du2020compositional}} is the first paper using EBMs for compositional visual generation. They propose three compositional operators for composing different concepts. Our work is inspired by \cite{du2020compositional}, but we compose diffusion models and achieve better results.

\subsection{Composing Language Descriptions}
\label{exp:compose_language}

Our approach can effectively compose natural language descriptions. 
We first show the image generation results of the pre-trained diffusion model, GLIDE \cite{nichol2021glide}, in \fig{fig:image_generation_glide}. 
We develop Composed GLIDE, a version of GLIDE that utilizes our compositional operators to combine textual descriptions, without further training. We compare this model to the original GLIDE model.

In \fig{fig:image_generation_glide}, GLIDE takes a single long sentence as input, for example, ``A pink sky in the horizon, a sailboat at the sea, and overwater bungalows''. In contrast, Composed GLIDE composes several short sentences using the concept conjunction operator, \eg ``A pink sky in the horizon'' AND ``A sailboat at the sea'' AND ``Overwater bungalows''.
While both GLIDE and Composed GLIDE can generate reasonable images containing objects described in the text prompt, our approach with the compositional operators can more accurately capture text details, such as the presence of ``a polar bear'' in the first example and the ``overwater bungalows'' in the third example.

\begin{figure}[t]
\begin{center}
\includegraphics[width=1\textwidth]{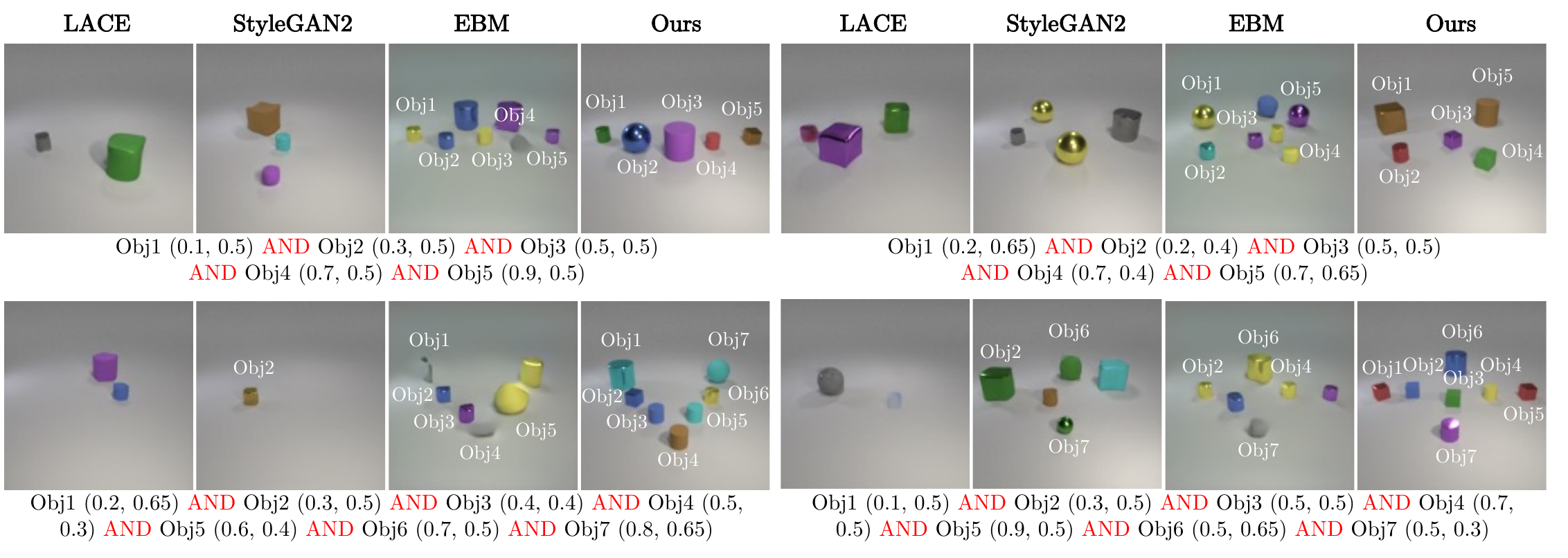}
\end{center}
\vspace{-15pt}
\caption{\small \textbf{Composing Objects.} Our method can compose multiple objects while baseline methods either miss objects or generate objects at wrong positions.}
\label{fig:image_generation_clevr}
\vspace{-15pt}
\end{figure}
\begin{table}[t]
    \centering
    \small
    \captionof{table}{\small Quantitative evaluation of $128\times128$ image generation results on CLEVR. The binary classification accuracy (Acc) and FID scores are reported. Our method outperforms baselines on all three test settings.}
    \vspace{10pt}
    \label{table:quant_clevr}
    \setlength{\tabcolsep}{1em}
    \scalebox{0.82}{
    \begin{tabular}{l|cc|cc|cc}
        \toprule
        \bf \multirow{2}{*}{Models} & \multicolumn{2}{c|}{\bf 1 Component} & \multicolumn{2}{c|}{\bf 2 Components} & \multicolumn{2}{c}{\bf 3 Components} \\
        & Acc (\%) $\uparrow$ & FID $\downarrow$ & Acc (\%) $\uparrow$ & FID $\downarrow$ & Acc (\%) $\uparrow$ & FID $\downarrow$ \\
        \midrule
        StyleGAN2-ADA~\cite{stylegan2ada} & 37.28 & 57.41 & - & - & - & - \\
        StyleGAN2~\cite{stylegan2} & 1.04 & 51.37 & 0.04 & 23.29 & 0.00 & 19.01 \\
        LACE~\cite{lace} & 0.70 & 50.92 & 0.00 &  22.83 & 0.00 & 19.62\\
        GLIDE~\cite{nichol2021glide} & 0.86 & 61.68 & 0.06 & 38.26 & 0.00 & 37.18 \\
        EBM~\cite{du2020compositional} & 70.54 & 78.63 & 28.22 & 65.45 & 7.34 & 58.33\\
        \bf Ours & \bf 86.42 & \bf 29.29 & \bf 59.20 & \bf 15.94 & \bf 31.36 & \bf 10.51 \\
        \bottomrule
    \end{tabular}}
    \vspace{-15pt}
\end{table}


\subsection{Composing Objects}
\label{exp:compose_object_attributes}

Given a set of 2D object positions, we aim to generate images containing objects at those positions.

\noindent\textbf{Qualitative results.}
We compare the proposed method and baselines on composing objects in \fig{fig:image_generation_clevr}. We only show the concept conjunction here because the object positions are not binary values, and thus negation of object positions is not interpretable. Given a set of object position labels, we compose them to generate images. Our model can generate images of objects at certain locations, while the baseline methods either miss objects or generate incorrect objects.

\noindent\textbf{Quantitative results.}
As shown in \tbl{table:quant_clevr}, our method outperforms baselines by a large margin. 
The binary classification accuracy of our method is $15.88\%$ higher than the best baseline, EBM, in the \emph{1 component} test setting and is $24.02\%$ higher than EBM in the more challenging \emph{3 Components} setting. 
Our method is more effective in zero-shot compositional generalization.
In addition, our method can generate images with lower FID scores, indicating the generated images are more similar to real images.

\begin{figure}[t]
\begin{center}
\includegraphics[width=0.9\textwidth]{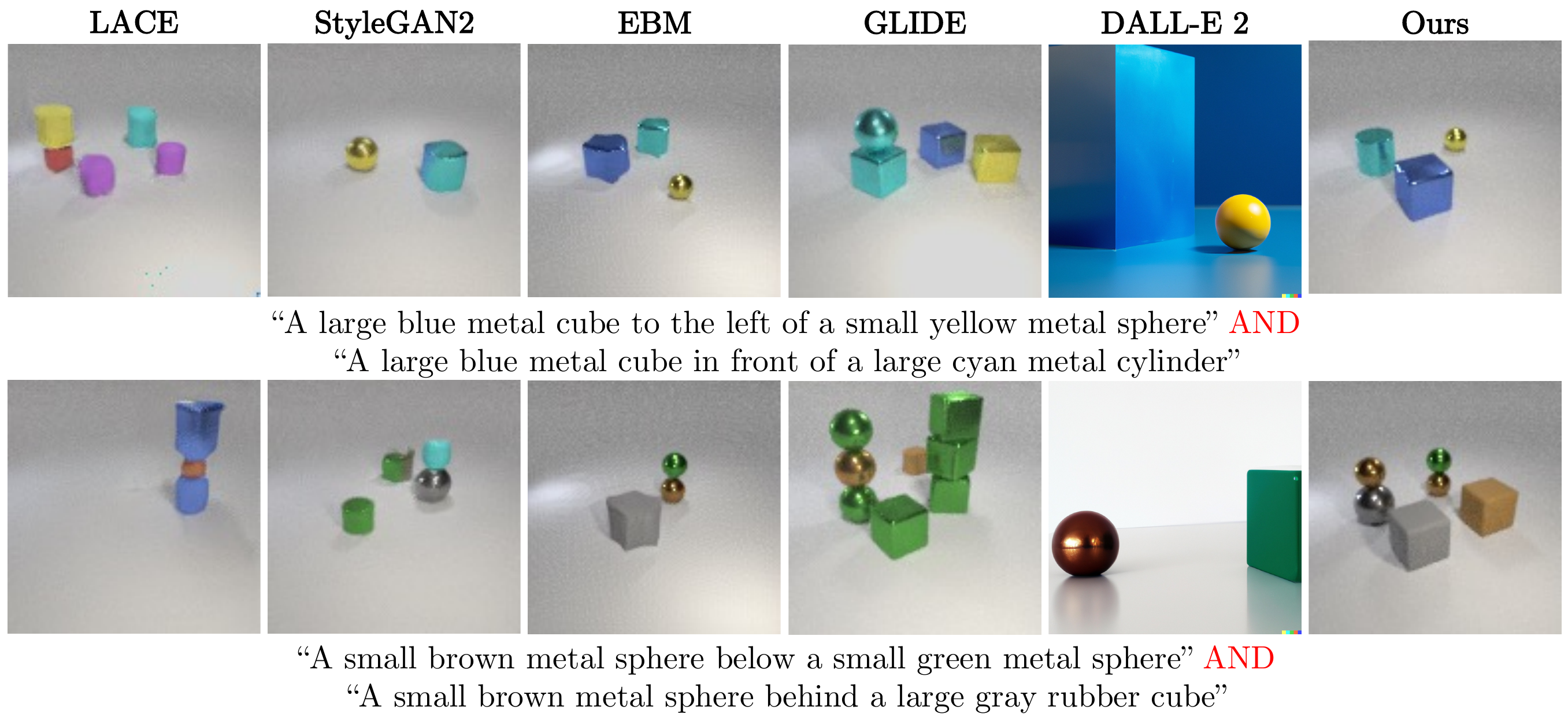}
\end{center}
\vspace{-15pt}
\caption{\small \textbf{Composing Visual Relations.} Image generation results on the Relational CLEVR dataset. 
Our model is trained to generate images conditioned on a single object relation, but during inference, our model can compose multiple object relations, generating better results than baselines.}
\label{fig:image_generation_clevr_rel}
\vspace{-15pt}
\end{figure}
\begin{table}[t]
    \centering
    \small
    \caption{{\small Quantitative evaluation of $128\times128$ image generation results on the Relational CLEVR dataset. The binary classification accuracy (Acc) and FID score on three test settings are reported. Although EBM performs well on binary classification accuracy, its FID score is much lower than other methods. Our method achieves comparable or better results than baselines.
    }}
    \vspace{5pt}
    \label{table:quant_clevr_rel}
    \setlength{\tabcolsep}{1em}
    \scalebox{0.82}{
    \begin{tabular}{l|cc|cc|cc}
        \toprule
        \bf \multirow{2}{*}{Models} & \multicolumn{2}{c|}{\bf 1 Component} & \multicolumn{2}{c|}{\bf 2 Components} & \multicolumn{2}{c}{\bf 3 Components} \\
        & Acc (\%) $\uparrow$ & FID $\downarrow$ & Acc (\%) $\uparrow$ & FID $\downarrow$ & Acc (\%) $\uparrow$ & FID $\downarrow$ \\
        \midrule
        StyleGAN2-ADA~\cite{stylegan2ada} & 67.71 & 20.55 & - & - & - & - \\
        StyleGAN2~\cite{stylegan2} & 20.18 & 22.29 & 1.66 & 30.58 & 0.16 & 31.30\\
        LACE~\cite{lace} & 1.10 & 40.54 & 0.10 & 40.61 & 0.04 & 40.60 \\ GLIDE~\cite{nichol2021glide} & 46.20 & \bf 17.61 & 8.86 & \bf 28.56 & 1.36 & 40.02 \\
        EBM~\cite{liu2021learning} & \bf 78.14 & 44.41 & \bf 24.16 & 55.89 & \bf 4.26 & 58.66\\
        \bf Ours & 60.40 & 29.06 & 21.84 & 29.82 & 2.80 & \bf 26.11 \\
        \bottomrule
    \end{tabular}}
    \vspace{-10pt}
\end{table}

\subsection{Composing Object Relations}
\label{exp:compose_object_relations}

\noindent\textbf{Qualitative results.}
We further compare the proposed approach and baselines on composing object relational descriptions in Figure~\ref{fig:image_generation_clevr_rel}. Our model is trained to generate images conditioned on a single object relation, but it can compose multiple object relations during inference without additional training. Both LACE and StyleGAN2 fail to capture object relations in the input sentences, but EBM and our method can correctly compose multiple object relations. 
Our method generates higher-quality images compared with EBM, \eg the object boundaries are sharper in our results than EBM. Surprisingly, DALL-E 2 and GLIDE can generate high-quality images, but they fail to understand object relations.

\noindent\textbf{Quantitative results.}
Similarly to experiments in \cref{exp:compose_object_attributes}, we evaluate the proposed method and baselines on three test settings in Table~\ref{table:quant_clevr_rel}. 
We train a binary classifier to evaluate whether an image contains objects that satisfy the input relational descriptions.
For binary classification accuracy, our method outperforms StyleGAN2, LACE, and GLIDE on all three test settings.
EBMs perform well on composing relational descriptions, but their FID scores are much worse than other methods, \ie their generated images are not realistic. StyleGAN2-ADA can obtain better accuracy and FID than our approach, but it cannot compose multiple concepts.

\subsection{Results analysis}
\label{exp:result_analysis}

\begin{figure}[t!]
  \begin{center}
    \includegraphics[width=1\textwidth]{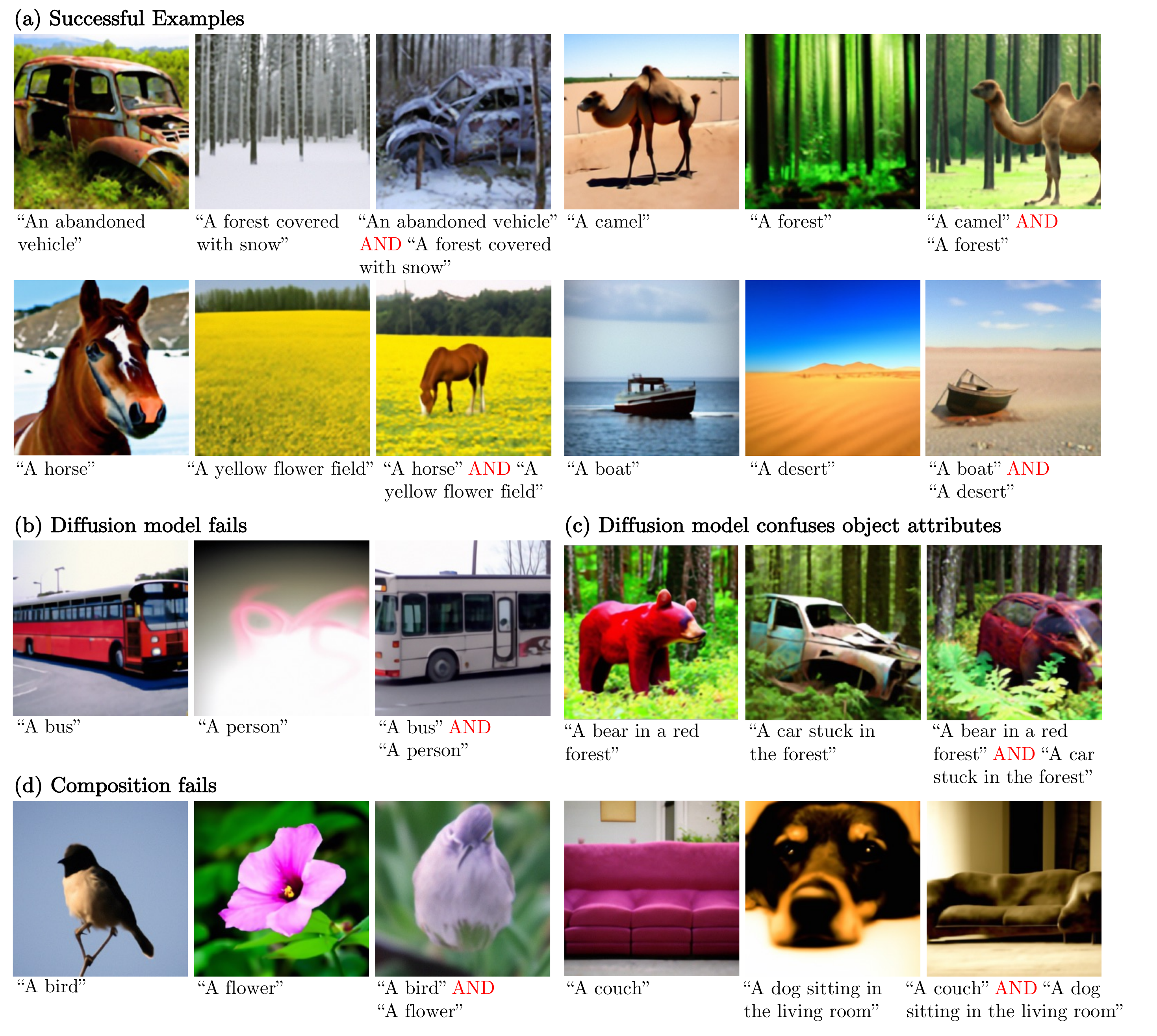}
  \end{center}
  \vspace{-15pt}
  \caption{\small \textbf{Qualitative results.} Successful examples (a) and failure examples (b-d) generated by the proposed method. There are three main types of failures: (b) The pre-trained diffusion model does not understand certain concepts, such as ``person''. (c) The pre-trained diffusion model confuses objects' attributes. (d) The composition fails. This usually happens when the objects are in the center of images.}
  \label{sec:result_analysis}
  \vspace{-15pt}
\end{figure}

We show our composed results on image generation and the results generated conditioned on each individual sentence description in \fig{sec:result_analysis}. 
We provide four successfully composed examples, where the generated images contain all the concepts mentioned in the input sentences.

\noindent\textbf{Failure cases}.
We observed three main failure cases of the proposed method. 
The first one is that the pre-trained diffusion models do not understand certain concepts, such as ``person'' in (b). This is because the pre-trained diffusion model, GLIDE~\cite{nichol2021glide}, is trained to avoid generating human images. The second type of failure is because the diffusion models confuse the objects' attributes. In (c), the generated image contains ``a red bear'' while the input is ``a bear in a red forest''.  The third type of failure is because the composition does not work, \eg the ``bird-shape and flower-color object'' and the ``dog-fur and sofa-shape object'' in (d). Such failures usually happen when the objects are in the center of the images.

\section{Conclusion}
In this paper, we compose diffusion models for image generation. 
By interpreting diffusion models as energy-based models, we may explicitly compose them and generate images with significantly more complex combinations that are never seen during training.
We propose two compositional operators, concept conjunction and negation, allowing us to compose diffusion models during the inference time without any additional training.
The proposed composable diffusion models can generate images conditioned on sentence descriptions, objects, object relations, and human facial attributes, and can generalize to new combinations that are rarely seen in the real world.
These results demonstrate the effectiveness of the proposed method for compositional visual generation.

A limitation of our current approach is that while we can compose multiple diffusion models together, they are instances of the same model. We found limited success when composing diffusion models trained on different datasets. In contrast, compositional generation with EBMs \cite{du2020compositional} can successfully compose multiple separately trained models. Incorporating additional structures into diffusion models from EBMs \cite{du2019implicit}, such as a conservative score field, can be a promising direction towards compositions of separately trained diffusion models.


\clearpage
%
%
\bibliographystyle{splncs04}
\bibliography{egbib}

\clearpage
\appendix
\appendix
\noindent\textbf{\Large{Appendix}}
\vspace{10pt}

In this appendix, we first demonstrate additional results in \cref{sup:additional_results}. We then show the details of training classifiers in \cref{sup:binary_classification}. In \cref{sup:our_details} and \cref{sup:baseline_details}, we show more details of our approach and baselines, respectively. Next, we provide the implementation details in \cref{sup:implementation}. Finally, we provide derivations of both the conjunction and negation operators in \cref{sup:proof}.

\section{Additional Results}
\label{sup:additional_results}

In this section, we first show the results of composing language descriptions to generate 3D meshes in \cref{exp:compose_lang_mesh}. We then show the results of composing human facial attributes in \cref{exp:compose_human}. Finally, we show more qualitative results in \cref{apx:more_qulitative_results}.

\subsection{Composing Language Descriptions for 3D Asset Generation}
\label{exp:compose_lang_mesh}

\begin{figure}[t]
\begin{center}
\includegraphics[width=1\textwidth]{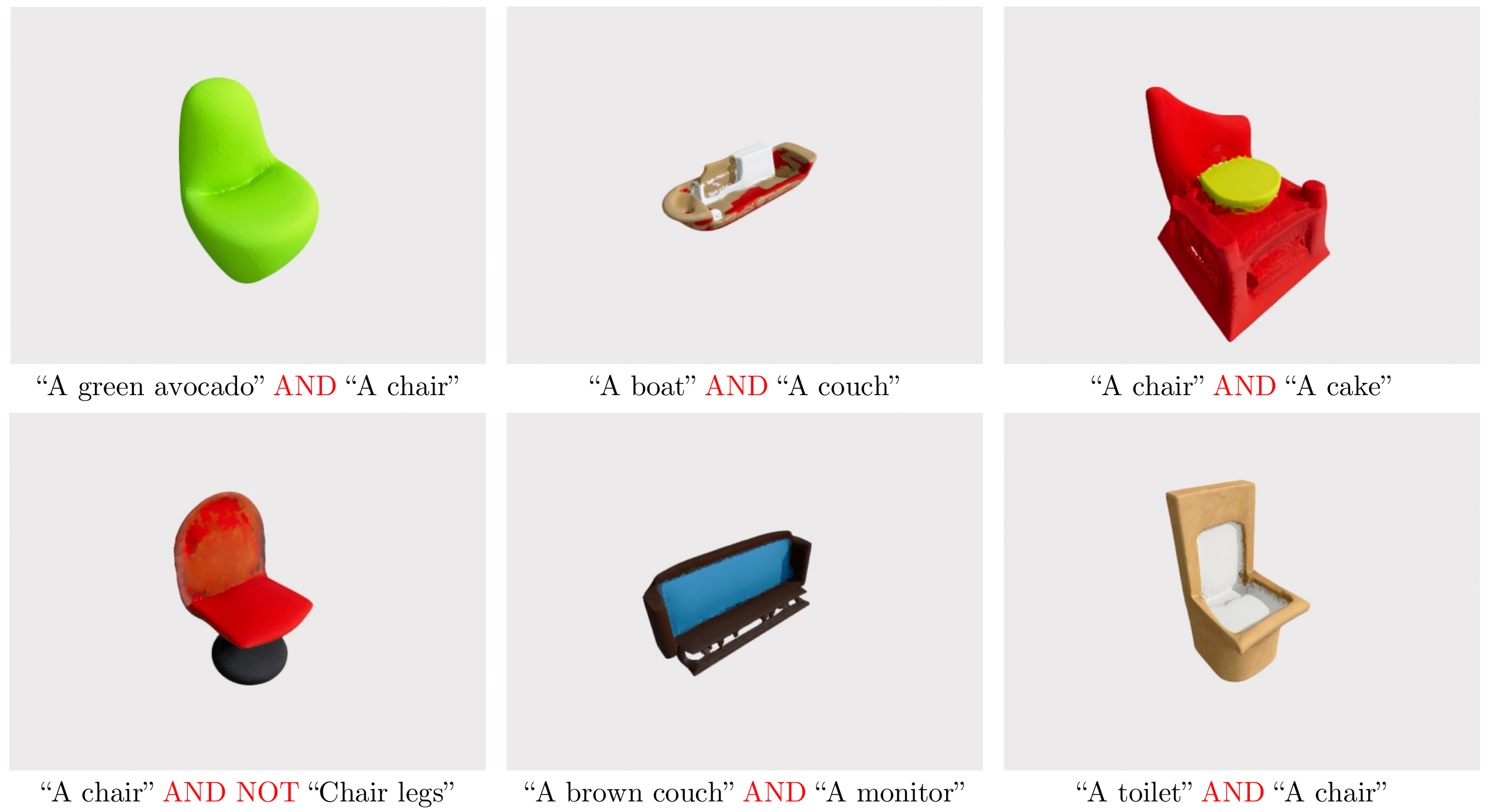}
\end{center}
\vspace{-15pt}
\caption{\small \textbf{Composing Language Descriptions for 3D Asset Generation}. We provide qualitative results of composing the pre-trained text-to-3D diffusion model,  Point-E~\cite{nichol2022point}, to generate interesting 3D hybrid objects.}
\label{fig:point-e}
\end{figure}

\noindent\textbf{Qualitative results.}
We demonstrate the proposed method of composing language descriptions for point cloud generation, which can be further used to generate 3D meshes.
We first use Point-E~\cite{nichol2022point}, the pre-trained 3D point cloud generation model, to generate the point clouds of an object based on the text description. We then convert the 3D point clouds into 3D meshes using marching cubes~\cite{lorensen1987marching}.
The results are shown in \fig{fig:point-e}. 


\subsection{Composing Human Facial Attributes}
\label{exp:compose_human}

\begin{figure}[h]
\begin{center}
\includegraphics[width=1\textwidth]{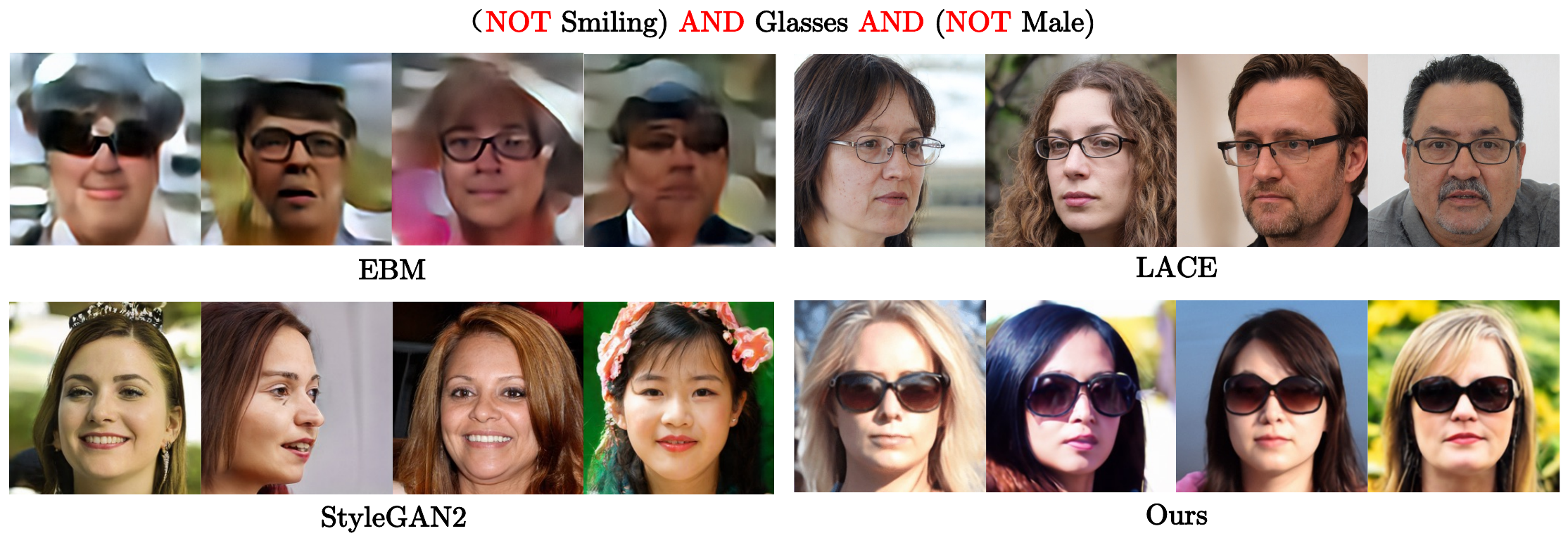}
\end{center}
\vspace{-15pt}
\caption{\small \textbf{Composing Facial Attributes.} Image generation results on the FFHQ dataset. 
Our model is trained to generate images conditioned on a single human facial attribute, but during inference, our model can recursively compose multiple facial attributes using the proposed compositional operators. The baselines either fail to compose attributes (StyleGAN2 and LACE) or generate low-quality images (EBM).}
\vspace{-10pt}
\label{fig:image_generation_ffhq}
\end{figure}

\noindent\textbf{Qualitative results.}
We compare the proposed method and baselines on composing facial attributes in \fig{fig:image_generation_ffhq}. 
We find that LACE and StyleGAN2 can generate high-fidelity images, but the generated images do not match the given labels. For example, StyleGAN2 generates humans without wearing glasses when the input label contains \emph{Glasses}, while LACE generates males sometimes when the input is \emph{Not Male}.
The image quality of EBM is much worse than other methods. In contrast, our method can generate high-fidelity images, containing all the attributes in the input label.

\noindent\textbf{Quantitative results.}
The results of our method and baselines on three test settings are shown in \tbl{table:quant_hhfq}. Our method is comparable with the best baseline on each evaluation metric.

\subsection{More Qualitative Results}
\label{apx:more_qulitative_results}
\begin{figure}[t]
\begin{center}
\includegraphics[width=1\textwidth]{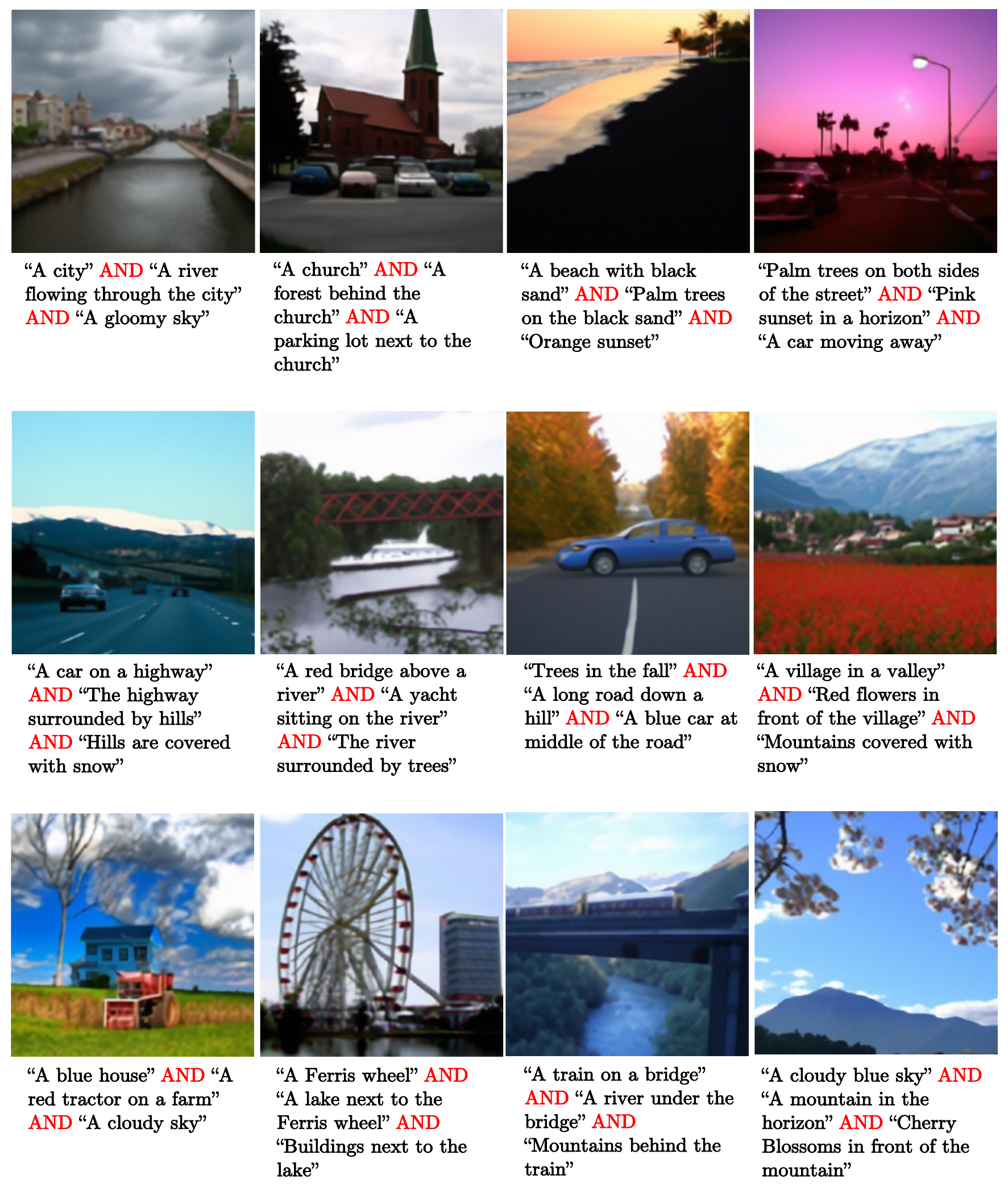}
\end{center}
\caption{\small \textbf{Composing Language Descriptions}. We provide more qualitative results of Composed GLIDE (Ours), a version of GLIDE~\cite{nichol2021glide} that utilizes our compositional operators to combine textual descriptions, without further training.}
\label{fig:image_generation_glide_sup}
\end{figure}

\begin{figure}[t]
\begin{center}
\includegraphics[width=1\textwidth]{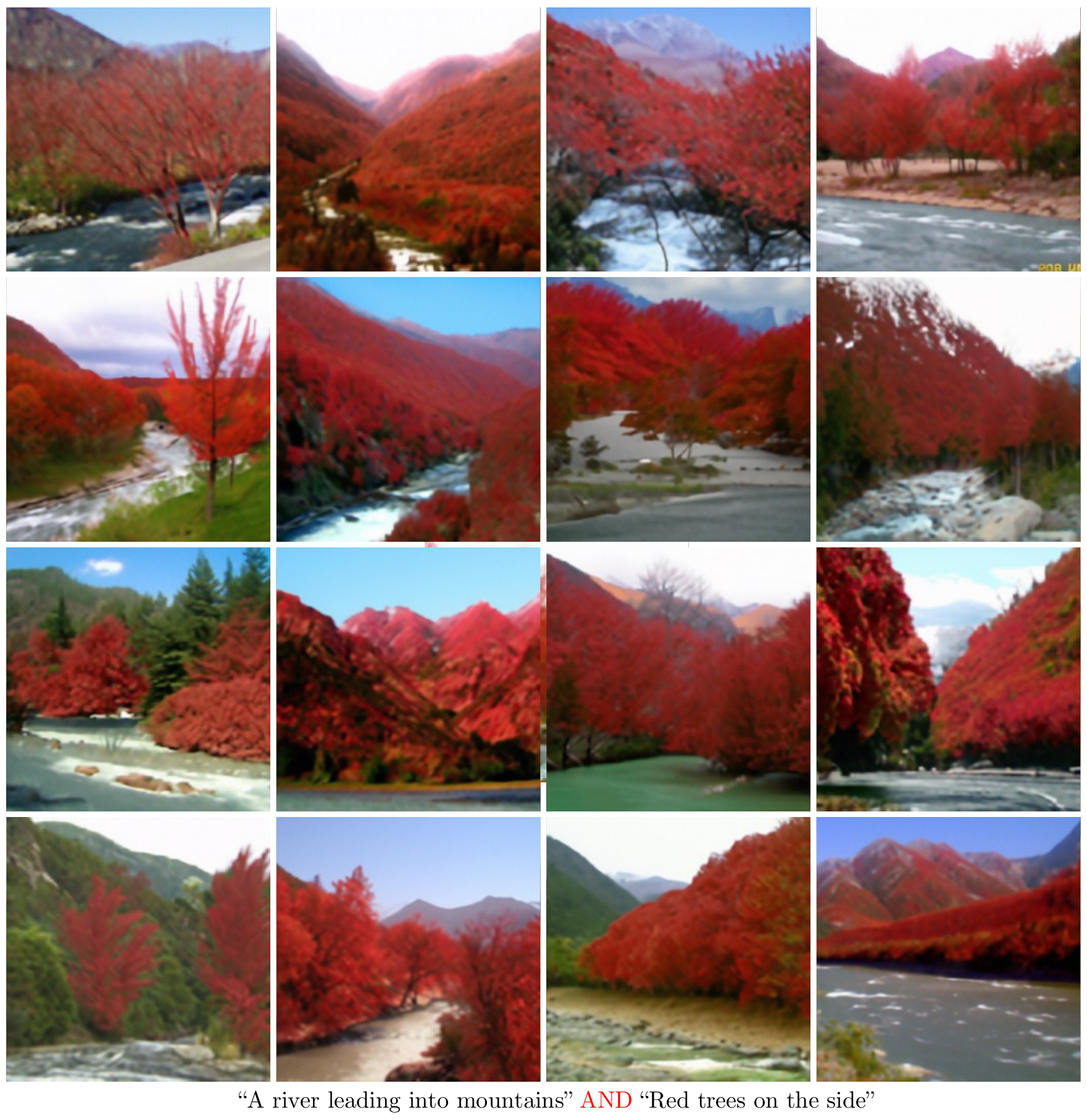}
\end{center}
\caption{\small \textbf{Composing Language Descriptions}. Images generated by our method, Composed GLIDE (Ours).}
\label{fig:image_generation_glide_grid_1}
\end{figure}

\begin{figure}[t]
\begin{center}
\includegraphics[width=1\textwidth]{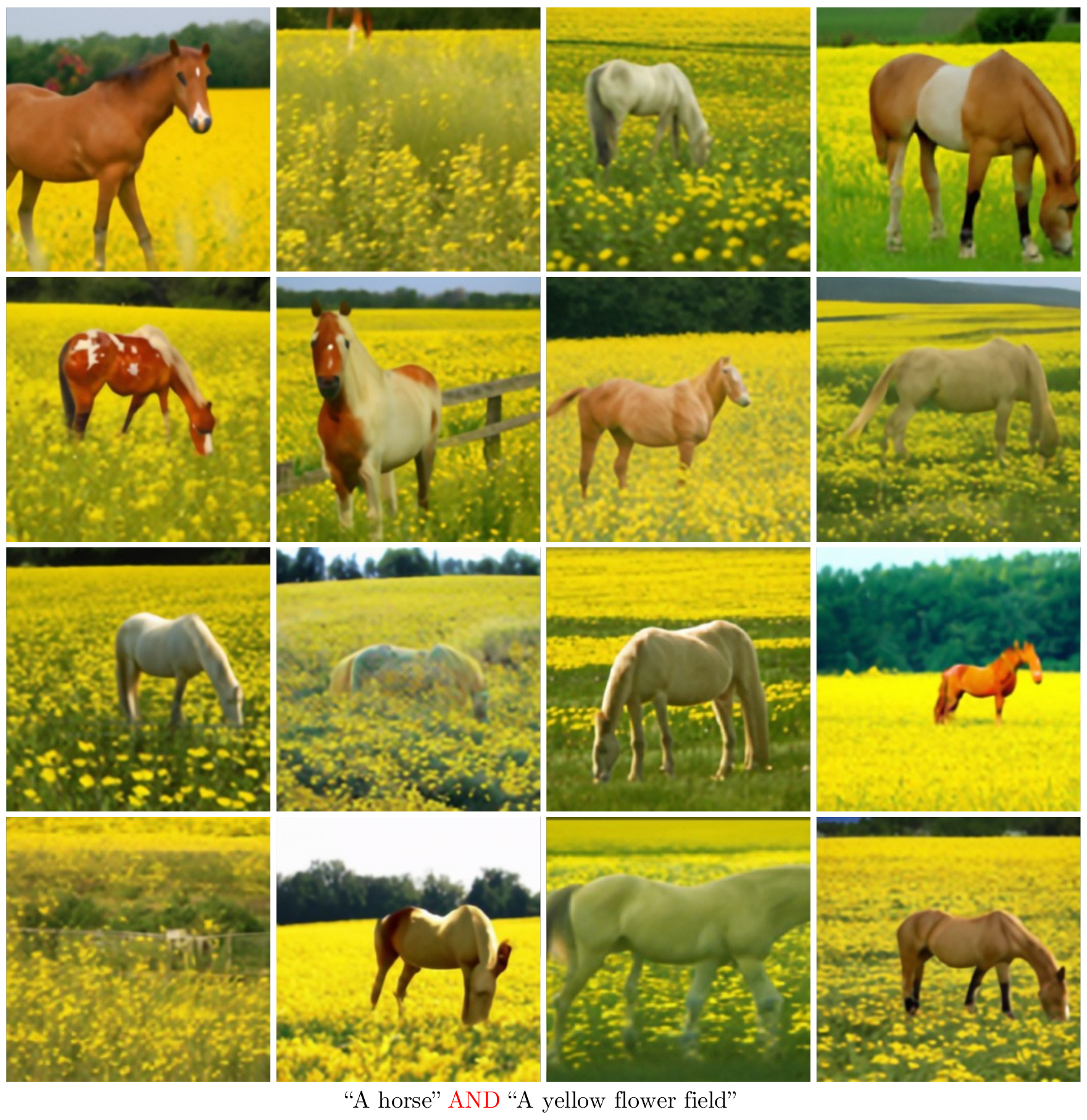}
\end{center}
\caption{\small \textbf{Composing Language Descriptions}. Images generated by our method, Composed GLIDE (Ours).}
\label{fig:image_generation_glide_grid_2}
\end{figure}

\begin{figure}[t]
\begin{center}
\includegraphics[width=1\textwidth]{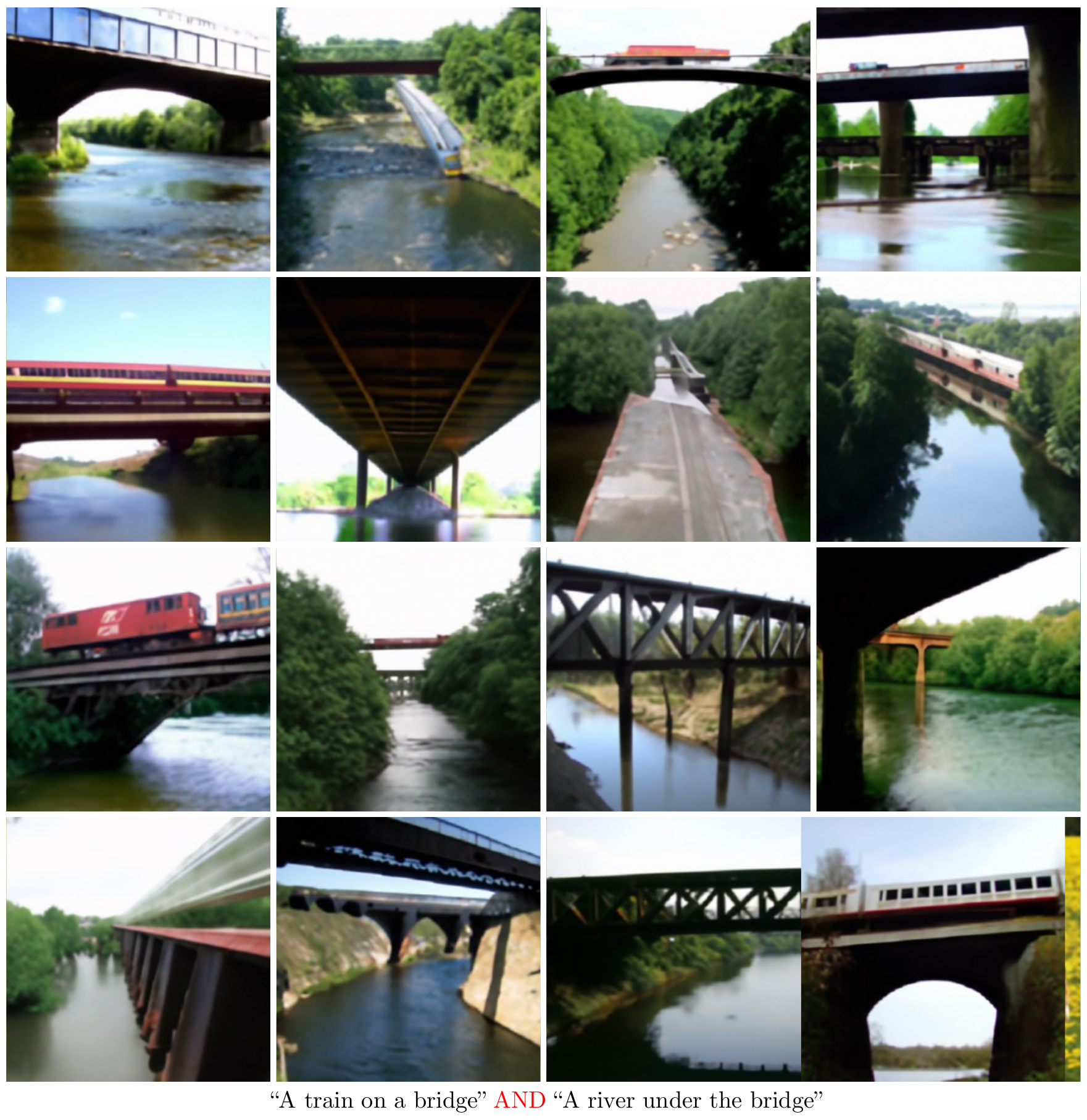}
\end{center}
\caption{\small \textbf{Composing Language Descriptions}. Images generated by our method, Composed GLIDE (Ours).}
\label{fig:image_generation_glide_grid_3}
\end{figure}

\begin{figure}[t]
\begin{center}
\includegraphics[width=1\textwidth]{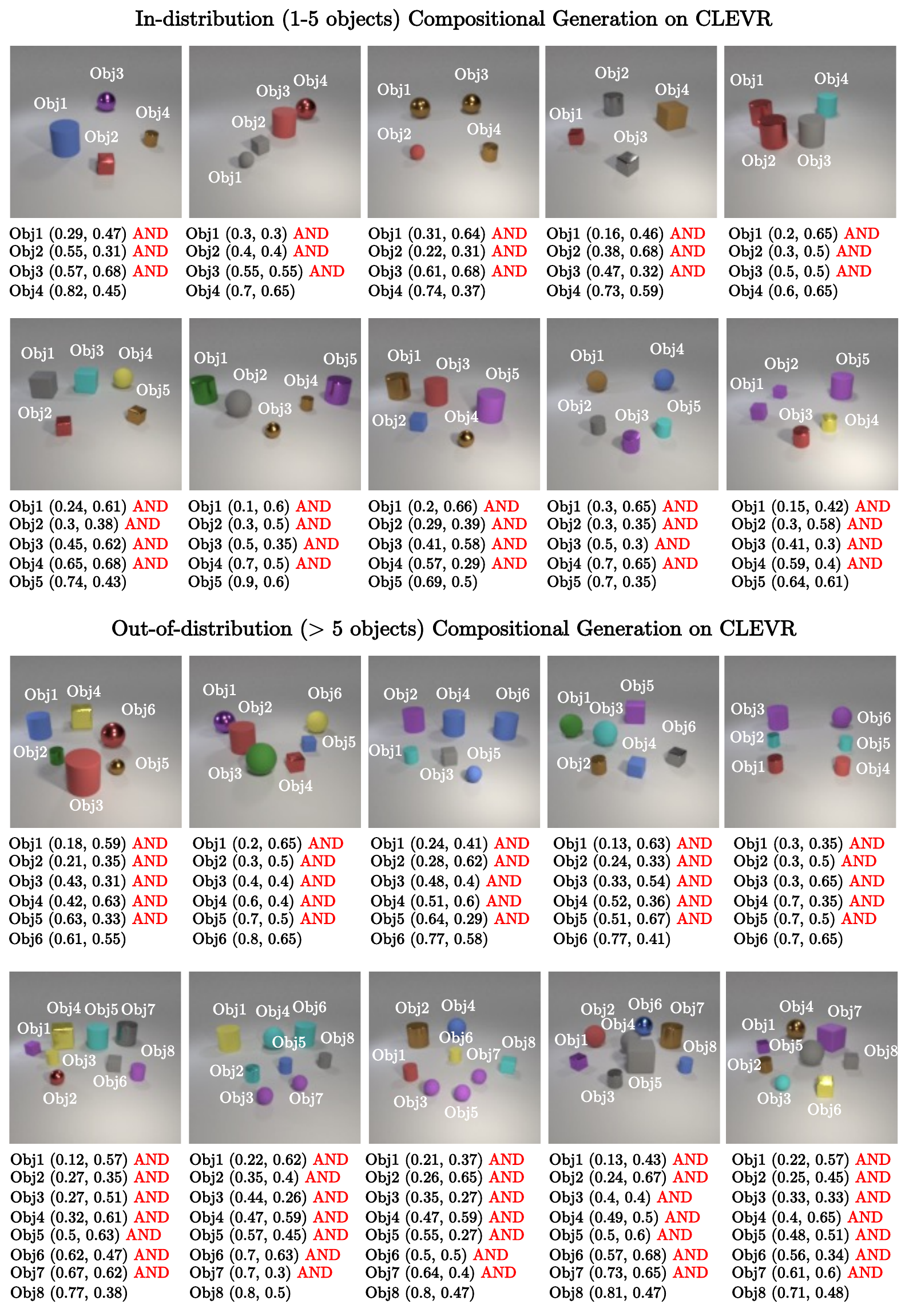}
\end{center}
\vspace{-10pt}
\caption{\small \textbf{Composing Objects.}
During inference, our model can generate images that contain multiple objects by composing their probability distributions using the conjunction operator. Note that the training set only contains images with fewer than five objects, but our model can compose more than five objects during inference.
}
\label{fig:image_generation_clevr_sup}
\end{figure}

We provide more qualitative results of the proposed method on composing concepts using the conjunction operator.
Figure \ref{fig:image_generation_glide_sup}, Figure \ref{fig:image_generation_glide_grid_1}, Figure \ref{fig:image_generation_glide_grid_2}, and Figure \ref{fig:image_generation_glide_grid_3} show more results of composing language descriptions. \fig{fig:image_generation_clevr_sup} shows additional results on composing objects on the CLEVR dataset. Our approach can reliably generate images conditioned on multiple concepts, even for combinations that are outside the training distribution.

\begin{figure}[t]
\begin{center}
\includegraphics[width=1\textwidth]{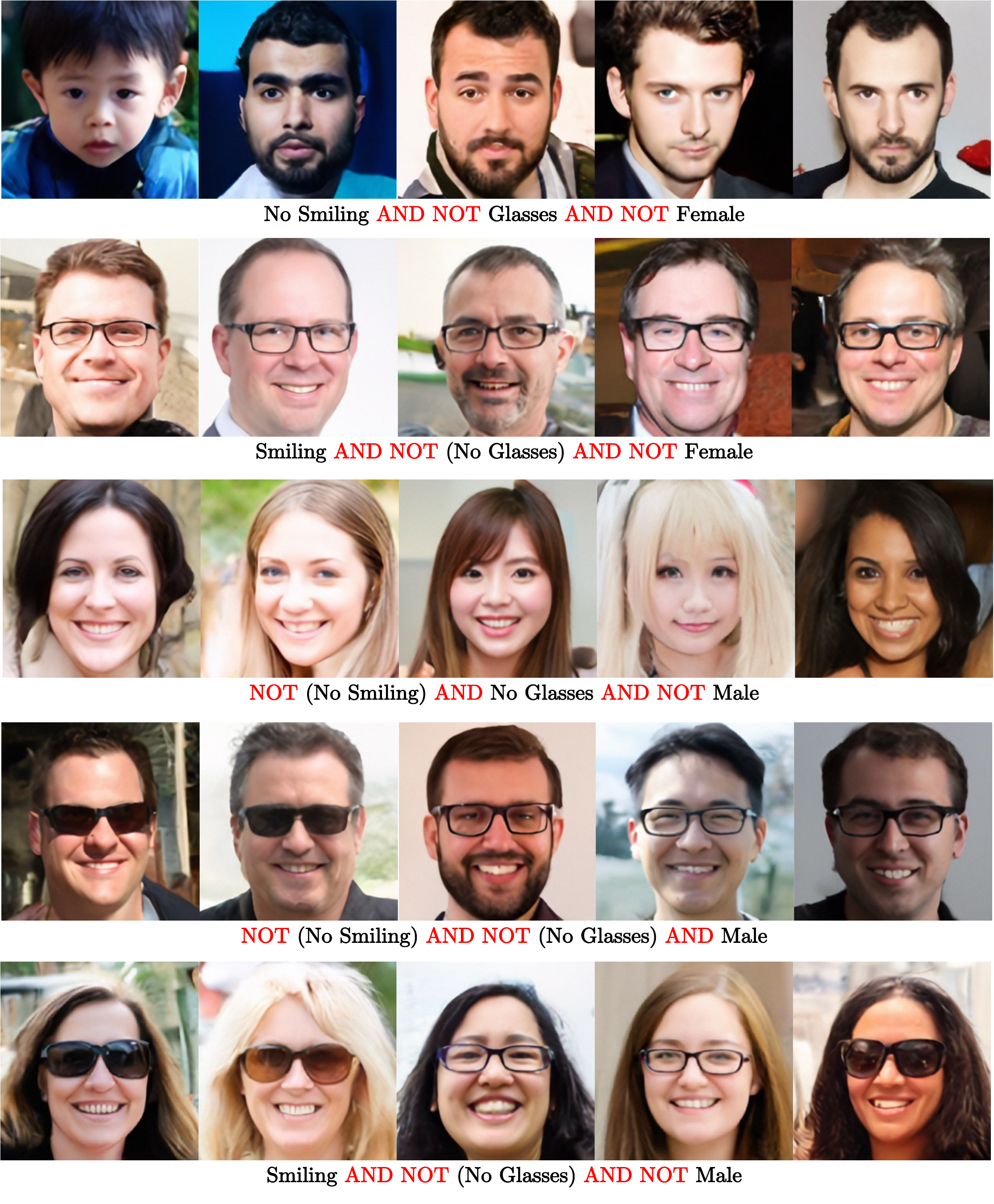}
\end{center}
\vspace{-10pt}
\caption{\small \textbf{Composing Human Facial Attributes.} During inference, our model can generate images that contain multiple attributes by composing their probability distributions using the conjunction and negation operators.}
\label{fig:image_generation_ffhq_sup}
\end{figure}

We further show the results of composing facial attributes on the FFHQ dataset in Figure \ref{fig:image_generation_ffhq_sup}. Our model is trained to generate images conditioned on a single human facial attribute, but it can compose multiple attributes during inference without further training by using the conjunction and negation compositional operators.
As shown in the fifth row of \fig{fig:image_generation_ffhq_sup}, our model can compose \emph{Not Male} and \emph{Glasses} and generate images with females wearing glasses. The proposed compositional operators allow our model to compose facial attributes recursively.

\begin{figure}
  \begin{center}
    \includegraphics[width=0.4\textwidth]{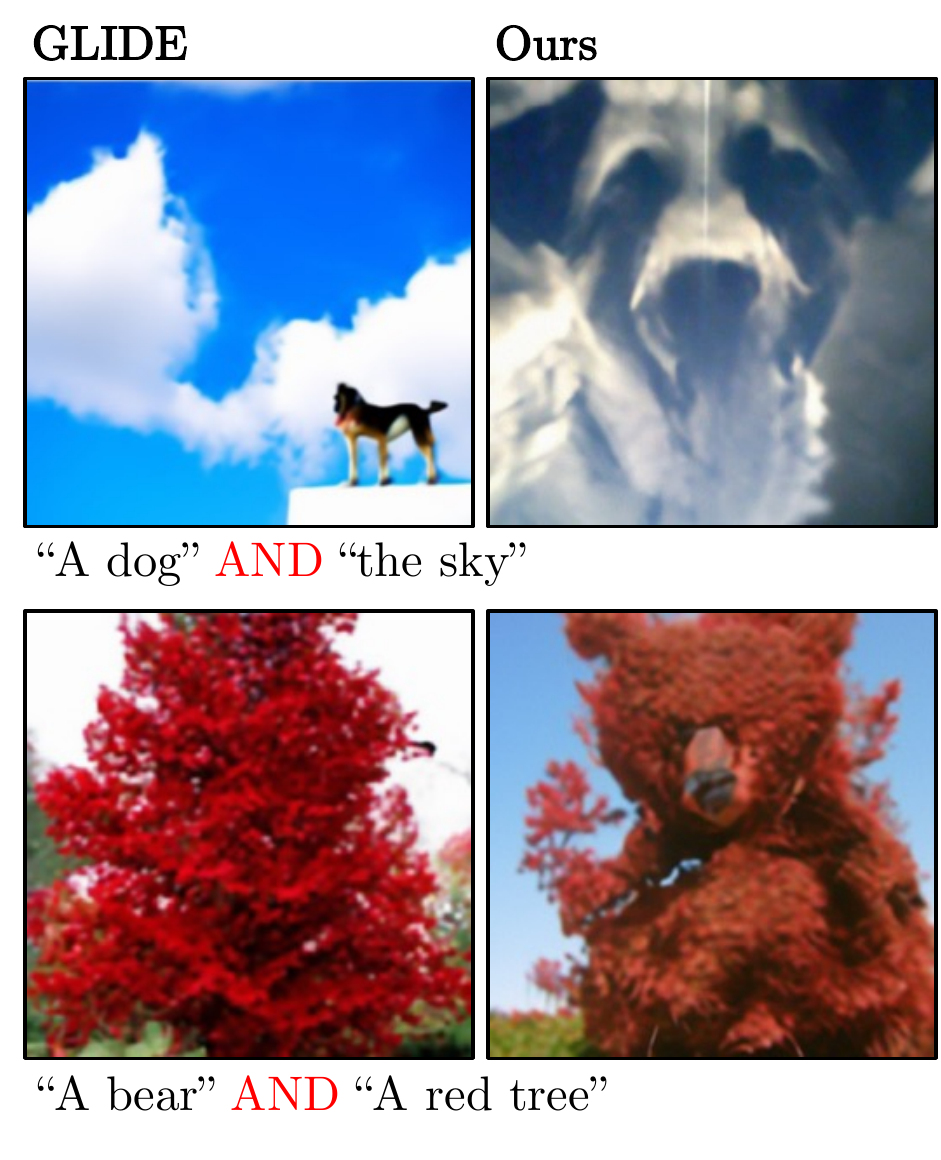}
  \end{center}
   \vspace{-20pt}
  \caption{\small Our method (composing multiple sentences) generates different styles of images compared to GLIDE (directly encodes the descriptions as a single long sentence).}
  \label{sec:interesting_cases}
\end{figure}

\textbf{Interesting cases}. 
As shown in \fig{sec:interesting_cases}, we find that our method, which combines multiple textual descriptions, can generate different styles of images compared to GLIDE, which directly encodes the descriptions as a single long sentence. Taking ``a dog'' and ``the sky'' as inputs, our method generates a dog-shaped cloud, whereas GLIDE generates a dog under the sky using the prompt ``a dog and the sky''.

\begin{table}[t]
    \centering
    \setlength{\tabcolsep}{1em}
    \small
    \caption{\small Image generation results on FFHQ. The binary classification accuracy (Acc) and FID are reported. Our method achieves comparable results with the best baseline on three test settings.}
    \vspace{10pt}
    \label{table:quant_hhfq}
    \scalebox{0.82}{
    \begin{tabular}{l|cc|cc|cc}
        \toprule
        \bf \multirow{2}{*}{Models} & \multicolumn{2}{c|}{\bf 1 Component} & \multicolumn{2}{c|}{\bf 2 Components} & \multicolumn{2}{c}{\bf 3 Components} \\
        & Acc ($\%$) $\uparrow$ & FID $\downarrow$ & Acc ($\%$) $\uparrow$ & FID $\downarrow$ & Acc ($\%$) $\uparrow$ & FID $\downarrow$ \\
        \midrule
        StyleGAN2-ADA~\cite{stylegan2ada} & 91.06 & \bf 10.75 & - & - & - & - \\
        StyleGAN2~\cite{stylegan2} & 58.90 & 18.04 & 30.68 &  18.06 & 16.96 & 18.06\\
        LACE~\cite{lace} & 97.60 & 28.21 & \bf 95.66 & 36.23 & \bf 80.88 & 34.64\\
        GLIDE~\cite{nichol2021glide} & 98.66 & 20.30 & 48.68 & 22.69 & 27.24 & 21.98 \\
        EBM~\cite{du2020compositional} & 98.74 & 89.95 & 93.10& 99.64 & 30.01 & 335.70\\
        \bf Ours & \bf 99.26 & 18.72 & 92.68 & \bf 17.22 & 68.86 & \bf 16.95 \\
        \bottomrule
    \end{tabular}}
    \vspace{-10pt}
\end{table}

\section{Details of Binary Classifiers}
\label{sup:binary_classification}
We provide more details of the binary classifiers in this section.

\noindent\textbf{CLEVR.}
The CLEVR dataset consists of 30,000 image-label pairs. We split the dataset into training and validation subsets. There are $24,000$ data pairs used for training and $6,000$ data pairs used for validation. 
We train a binary classifier to evaluate whether there is an object appearing at a particular position of an image.
The classifier achieves an accuracy of $99.05\%$ on the validation set, which is used to evaluate the quality of generated images.

\noindent\textbf{Relational CLEVR.}
The Relational CLEVR~\cite{liu2021learning} dataset contains $50,000$ images at $128 \times 128$ resolution. We split the dataset into $40,000$ training data and $10,000$ validation data. Then we train a binary classifier to evaluate whether an image contains an object relational description. The trained classifier achieves an accuracy of $99.80\%$ on the validation set.

\noindent\textbf{FFHQ.}
We use $30,000$ image-label pairs from CelebA-HQ \cite{karras2017progressive} to train a classifier to evaluate the generated images. We split the dataset into the training ($24,000$ data pairs) and validation ($6,000$ data pairs) subsets.
We select three attributes (\ie \emph{smiling}, \emph{glasses}, and \emph{gender}) to evaluate the compositional ability of our approach and baselines. We thus train three binary classifiers to evaluate the \emph{smiling}, \emph{glasses}, and \emph{gender} concepts, respectively. Our classifiers achieve $95.01\%$, $99.20\%$ and $97.49\%$ accuracy on the validation sets of \emph{smiling}, \emph{glasses}, and \emph{gender}, respectively.



\section{Details of Our Approach}
\label{sup:our_details}

\noindent\textbf{Training.}
Our approach is implemented based on the code from \cite{nichol2021improved,nichol2021glide}.
Ho \etal~\cite{ho2021classifier} introduce a technique to train the conditional and unconditional diffusion models at the same time by masking some labels as nulls. We use the same approach to train diffusion models. For each data point, its label has a $10\%$ chance of being replaced by a null label which is used to estimate the unconditional score. 

\noindent\textbf{Inference.}
To generate FFHQ images, we first generate images at $64\times64$ resolution and then upsample the images to $256\times256$ using a sampler provided by~\cite{nichol2021glide}.
For CLEVR images, we generate images at $128\times128$ resolution directly.

\noindent\textbf{Label Encoding.}
On the FFHQ dataset, we use three human facial attributes, \ie \emph{smile}, \emph{glasses} and \emph{gender}.
For the \emph{smile} and \emph{glasses} attributes, label $1$ indicates an image containing the attribute; otherwise, the label is $0$.
For the \emph{gender} attribute, label $0$ indicates ``male'', while label $1$ represents ``female''. 
We use the embedding layer $nn.Embedding(7, d)$ to encode the attribute labels. The first six dimensions represent the attribute labels and the last dimension indicates the null class. The labels are encoded as a $d$-dimension feature vector, which is then fused with the time embedding to estimate the score $\epsilon_{\theta}$.

On the CLEVR dataset, we encode the $(x,y)$ coordinates using a linear layer $nn.Linear(2, d)$, where $d$ is the dimension of the output feature.
The coordinate embedding is then fused with the time embedding to estimate the score $\epsilon_{\theta}$.

\vspace{-5pt}
\section{Details of Baselines}
\label{sup:baseline_details}
\vspace{-5pt}

\noindent\textbf{StyleGAN2-ADA.}
\label{baselien:stylegan2-ada}
On each dataset, we train a conditional StyleGAN2-ADA model using the ``stylegan'' configuration provided by~\cite{stylegan2ada} without using augmentations. 

\noindent\textbf{StyleGAN2.}
\label{baselien:stylegan2}
We use the pre-trained StyleGAN2 model~\cite{stylegan2} to evaluate its performance on facial image generation. 
As there is no pre-trained model for object generation, we use the same code to train a model on the CLEVR dataset for image generation conditioned on object positions.
We use the ``config-f'' setting provided by~\cite{stylegan2}. To enable image generation conditioned on multiple concepts, we train a binary classifier on each task. During inference, we optimize the latent code z by decreasing the binary classification loss of the generated image and the given label. We use the resultant latent code to generate images.

\noindent\textbf{LACE.} 
LACE~\cite{lace} trains classifiers for image generation using the generated images from StyleGAN2 and labels provided by the neural network. For the CLEVR dataset, we first generate $10,000$ images using the same StyleGAN2 model that was trained on CLEVR in  Section~\ref{baselien:stylegan2}. Then we modify the code to train a position annotator using a DenseNet~\cite{huang2017densely} model provided by LACE to label the object positions of generated images. Lastly, we train a classifier conditioned on object coordinates using the code provided by \cite{lace}. For FFHQ, we use the off-the-shelf pre-trained model from \cite{lace} for comparison. 

\noindent\textbf{GLIDE.} We use the small GLIDE model released by \cite{nichol2021glide} in our experiments. 
We develop Composed GLIDE (Ours), a version of GLIDE that utilizes our compositional operators to combine textual descriptions, without further training.
We compare it to the original GLIDE, which
directly encodes the descriptions as a single long sentence.
\cite{nichol2021glide} also releases an upsampler model to upsample the generated images from a resolution of $64 \times 64$ to a resolution of $256 \times 256$.
We use the upsampler model for both the GLIDE and Composed GLIDE (Ours).

\noindent\textbf{Energy-based models (EBMs).}
We train energy-based models using the codebase from~\cite{du2020improved}, where we encode discrete labels and continuous labels using an embedding layer and a linear layer, respectively. We use the inference code from~\cite{du2020compositional} to compose multiple concepts.

\section{Implementation Details}
\label{sup:implementation}

Each model is trained on a single Tesla V100 32GB GPU.

\noindent\textbf{StyleGAN2-ADA.}
\label{impl:stylegan2-ada}
Each conditional StyleGAN2-ADA model is trained for two days. We use the Adam optimizer~\cite{kingma2014adam} with $\beta_1 = 0$ and $\beta_2 = 0.99$
to train the models.

\noindent\textbf{StyleGAN2.}
\label{impl:stylegan2}
We train a StyleGAN2 model for two days on both CLEVR and Relational CLEVR datasets. We use the Adam optimizer~\cite{kingma2014adam} with $\beta_1 = 0$ and $\beta_2 = 0.99$ to train the StyleGAN2 models. 
It takes 2 hours to train a binary classifier.
The classifiers are trained using the Adam optimizer with $\beta_1 = 0$ and $\beta_2 = 0.99$.
For the FFHQ dataset, We use the pre-trained model provided by \cite{stylegan2}.

\noindent\textbf{LACE.}
LACE uses the pre-trained model provided by \cite{stylegan2} on the FFHQ dataset.
For both CLEVR and Relational CLEVR datasets, we directly reuse the trained StyleGAN2 model as described in Section~\ref{impl:stylegan2}. 
It takes less than $10$ minutes to train the classifier on each dataset.

\noindent\textbf{EBMs.}
In our experiments, we use the same setting to train models on different datasets. We use the Adam optimizer~\cite{kingma2014adam} with a learning rate of $10^{-4}$. For MCMC sampling, we use a step size of 300 and 80 iterations. Similarly, the model is trained for two days on each dataset.

\noindent\textbf{Ours.}
To train diffusion models on CLEVR and FFHQ, we use $1,000$ diffusion steps, and the cosine noise schedule. We use the AdamW optimizer~\cite{loshchilov2017decoupled} with $\beta_1 = 0.9$ and $\beta_2 = 0.999$. 
We train the diffusion models on CLEVR for seven days ($750,000$ iterations) and FFHQ for two days ($250,000$ iterations).

\section{Derivation}
\label{sup:proof}
\subsection{Conjunction Operator (AND)}
Given a set of independent concepts $\{\vc_1, \vc_2, \ldots, \vc_n\}$, the joint probability distribution can be factorized as follows:

\begin{equation}
    \begin{split}
        p(\vx |\vc_1, \ldots, \vc_n) \propto p(\vx, \vc_1, \ldots, \vc_n) = p(\vx) \prod_{i=1}^n p(\vc_i|\vx)
    \end{split}
\end{equation}
We can rewrite above expression using $p(\vc_i | \vx) \propto \frac{p(\vx | \vc_i)}{p(\vx)}$:

\begin{equation}
    p(\vx) \prod_{i=1}^n p(\vc_i|\vx) \propto p(\vx) \prod_{i=1}^n \frac{p(\vx | \vc_i)}{p(\vx)}
\end{equation}
Then we take a gradient of logarithm on both sides w.r.t $\vx$:
\begin{equation}
    \begin{split}
        \nabla_{\vx} \log p(\vx |\vc_1, \ldots, \vc_n) &= \nabla_{\vx}\log p(\vx) + \sum_{i=1}^n \bigl(\nabla_{\vx}\log p(\vx | \vc_i) - \nabla_{\vx}\log p(\vx) \bigl) \\
        &= \epsilon_\theta(\vx_t, t) + \sum_{i=1}^n \bigl(\epsilon_\theta(\vx_t, t | \vc_i) - \epsilon_\theta(\vx_t, t) \bigl)
    \end{split}
\end{equation}
Finally, we may obtain a modified score prediction from the above expression $\hat{\epsilon}_\theta(\vx_t, t |\vc_1, \ldots, \vc_n)$, where $w_i$ controls the temperature of each implicit classifier: 
\begin{equation}
    \hat{\epsilon}_{\theta}(\vx_t, t |\vc_1, \ldots, \vc_n) = \epsilon_\theta(\vx_t, t) + \sum_{i=1}^n w_i\bigl(\epsilon_\theta(\vx_t, t | \vc_i) - \epsilon_\theta(\vx_t, t) \bigl)
\end{equation}

\noindent In the setting where only one concept $\vc_1$ is conditioned for sampling, the above equation will reduce to classifier-free guidance~\cite{ho2021classifier}:
\begin{equation}
    \hat{\epsilon}_\theta(\vx_t, t | \vc_1) = \epsilon_\theta(\vx_t, t) + w \bigl(\epsilon_\theta(\vx_t, t| \vc_1) - \epsilon_{\theta}(\vx_t, t)\bigl),
\end{equation}
where the temperature scaling $w > 1$.

\subsection{Negation Operator (NOT)}
Given two independent concepts $\{\vc_1, \vc_2\}$, the joint probability distribution where we negate the concept $\vc_1$ can be similarly written as:

\begin{equation}
    p(\vx | \text{not } \vc_1, \vc_2) \propto p(\vx, \text{not }\vc_1, \vc_2) \propto p(\vx)\frac{p(\vc_2 | \vx)}{p(\vc_1 | \vx)} \propto p(\vx)\frac{p(\vx | \vc_2)}{p(\vx | \vc_1)}
\end{equation}
Then we take a gradient of logarithm on both sides w.r.t $\vx$ as follows:
\begin{equation}
    \begin{split}
        \nabla_{\vx} \log p(\vx | \text{not } \vc_1, \vc_2) &= \nabla_{\vx} \log p(\vx) + \nabla_{\vx} \log p(\vx | \vc_2) - \nabla_{\vx} \log p(\vx | \vc_1) \\
        &= \epsilon_\theta(\vx_t, t) + \epsilon_\theta(\vx_t, t | \vc_2) - \epsilon_\theta(\vx_t, t | \vc_1)
    \end{split}
\end{equation}
Finally, we may obtain a modified score prediction from the above, where $w$ is a tunable coefficient that determines the weight of the negation: 
\begin{equation}
    \hat{\epsilon}_{\theta}(\vx_t, t |\text{not }\vc_1, \vc_2) = \epsilon_\theta(\vx_t, t) + w \bigl(\epsilon_\theta(\vx_t, t | \vc_2) - \epsilon_\theta(\vx_t, t | \vc_1) \bigl)
\end{equation}

\end{document}